%% file: paper.tex
\documentclass[sigconf]{acmart}
\usepackage{enumitem}
\usepackage{flushend}
\usepackage{balance}
\usepackage{booktabs}
\usepackage{todonotes}
\usepackage{epsfig}
\usepackage{graphicx}
\usepackage{amsmath}
\usepackage{amssymb}
\usepackage{lipsum}
\usepackage{subfig}
\usepackage{color}

\usepackage[normalem]{ulem}
\usepackage{bigstrut}
\usepackage{multirow}
\usepackage{siunitx}
\usepackage{ulem}
\usepackage{listings}
\usepackage{eso-pic}
\usepackage{tikz}
\usepackage{xspace}
\usepackage{fancyhdr}

\usepackage[noend]{algpseudocode}
\usepackage[ruled,vlined,algo2e]{algorithm2e}
\usepackage{stfloats}
\usepackage[compact]{titlesec}

\usepackage[bookmarks=true,breaklinks=true,letterpaper=true,colorlinks,linkcolor=blue,citecolor=blue,urlcolor=black]{hyperref}

\input{macros}
\graphicspath{{figs/}}

\AtBeginDocument{%
  \providecommand\BibTeX{{%
    \normalfont B\kern-0.5em{\scshape i\kern-0.25em b}\kern-0.8em\TeX}}}

\setcopyright{acmcopyright}

\begin{document}

\copyrightyear{2019} 
\acmYear{2019} 
\acmConference[MICRO-52]{The 52nd Annual IEEE/ACM International Symposium on Microarchitecture}{October 12--16, 2019}{Columbus, OH, USA}
\acmBooktitle{The 52nd Annual IEEE/ACM International Symposium on Microarchitecture (MICRO-52), October 12--16, 2019, Columbus, OH, USA}
\acmPrice{15.00}
\acmDOI{10.1145/3352460.3358259}
\acmISBN{978-1-4503-6938-1/19/10}

\title{Tigris: Architecture and Algorithms for 3D Perception \\ in Point Clouds}

\author{Tiancheng Xu}
\authornote{Tiancheng Xu and Boyuan Tian are co-primary authors.}
\email{txu17@ur.rochester.edu}
\affiliation{}

\author{Boyuan Tian}
\authornotemark[1]
\email{btian2@ur.rochester.edu}
\affiliation{}

\author{Yuhao Zhu}
\email{yzhu@rochester.edu}
\affiliation{}

\author{}
\affiliation{
\institution{Department of Computer Science University of Rochester\\\texttt{http://horizon-lab.org}}\vspace{20pt}}

\renewcommand{\shortauthors}{Tiancheng Xu, Boyuan Tian, Yuhao Zhu}
\renewcommand{\shorttitle}{Tigris: Architecture and Algorithms for 3D Perception in Point Clouds}

\input{abst}



\artifacts{Configurable Point Cloud Pipeline: \url{https://github.com/horizon-research/pointcloud-pipeline}}

\setlength{\textfloatsep}{6pt}
\setlength{\floatsep}{6pt}

\titlespacing*{\section}{0pt}{8pt plus 0pt minus 0pt}{4pt plus 0pt minus 0pt}
\titlespacing*{\subsection}{0pt}{6pt plus 0pt minus 0pt}{3pt plus 0pt minus 0pt}

\maketitle

\input{intro}

\input{background}

\input{char}

\input{algo}

\input{arch}

\input{eval}

\input{related}
\input{conc}

\raggedright
\balance
\Urlmuskip = 0mu plus 1mu\relax
\bibliographystyle{ACM-Reference-Format}
\bibliography{refs}

\end{document}

%% file: macros.tex
\ifx\figurename\undefined \def\figurename{Figure}\fi
\renewcommand{\figurename}{Fig.}
\renewcommand{\paragraph}[1]{\textbf{#1}~~}

\newcommand{\Sect}[1]{Sec.~\ref{#1}}
\newcommand{\Fig}[1]{Fig.~\ref{#1}}
\newcommand{\Tbl}[1]{Tbl.~\ref{#1}}

\newcommand{\Alg}[1]{Algo.~\ref{#1}}

\newcommand{\specialcell}[2][c]{\begin{tabular}[#1]{@{}c@{}}#2\end{tabular}}

\newcommand{\proj}{\textsc{Tigris}\xspace}

\newcommand{\sys}[1]{\textsc{#1}\xspace}

\newcommand{\RNum}[1]{\uppercase\expandafter{\romannumeral #1\relax}}


%% file: abst.tex
\begin{abstract}

Machine perception applications are increasingly moving toward manipulating and processing 3D point cloud. This paper focuses on point cloud \textit{registration}, a key primitive of 3D data processing widely used in high-level tasks such as odometry, simultaneous localization and mapping, and 3D reconstruction. As these applications are routinely deployed in energy-constrained environments, real-time and energy-efficient point cloud registration is critical.

We present \proj, an algorithm-architecture co-designed system specialized for point cloud registration. Through an extensive exploration of the registration pipeline design space, we find that, while different design points make vastly different trade-offs between accuracy and performance, KD-tree search is a common performance bottleneck, and thus is an ideal candidate for architectural specialization. While KD-tree search is inherently sequential, we propose an acceleration-amenable data structure and search algorithm that exposes different forms of parallelism of KD-tree search in the context of point cloud registration. The co-designed accelerator systematically exploits the parallelism while incorporating a set of architectural techniques that further improve the accelerator efficiency. Overall, \proj achieves 77.2$\times$ speedup and 7.4$\times$ power reduction in KD-tree search over an RTX 2080 Ti GPU, which translates to a 41.7\% registration performance improvements and 3.0$\times$ power reduction.
\end{abstract}

%% file: intro.tex
\section{Introduction}
\label{sec:intro}

Enabling machines to perceive, process, and understand visual data plays a vital role toward the promise of an intelligent future. While traditional machine perception focuses mostly on  processing 2D visual data such as images and videos, 3D data -- represented using point cloud -- that provides a three-dimensional measure of object shapes has become increasingly important. The proliferation of 3D data acquisition systems such as LiDAR, time-of-flight cameras, and structured-light scanners stimulates the development of point cloud processing algorithms. As a result, point cloud-based algorithms have become central to many application domains ranging from robotics navigation~\cite{whitty2010autonomous}, Augmented and Virtual reality~\cite{stets2017visualization}, to 3D reconstruction~\cite{vosselman20013d}.



\begin{figure}[h]
	\vspace{-5pt}
	\centering 
	\subfloat[Data frame A]{
	  \label{fig:reg_demo_left}
	  \includegraphics[width=0.3\linewidth]{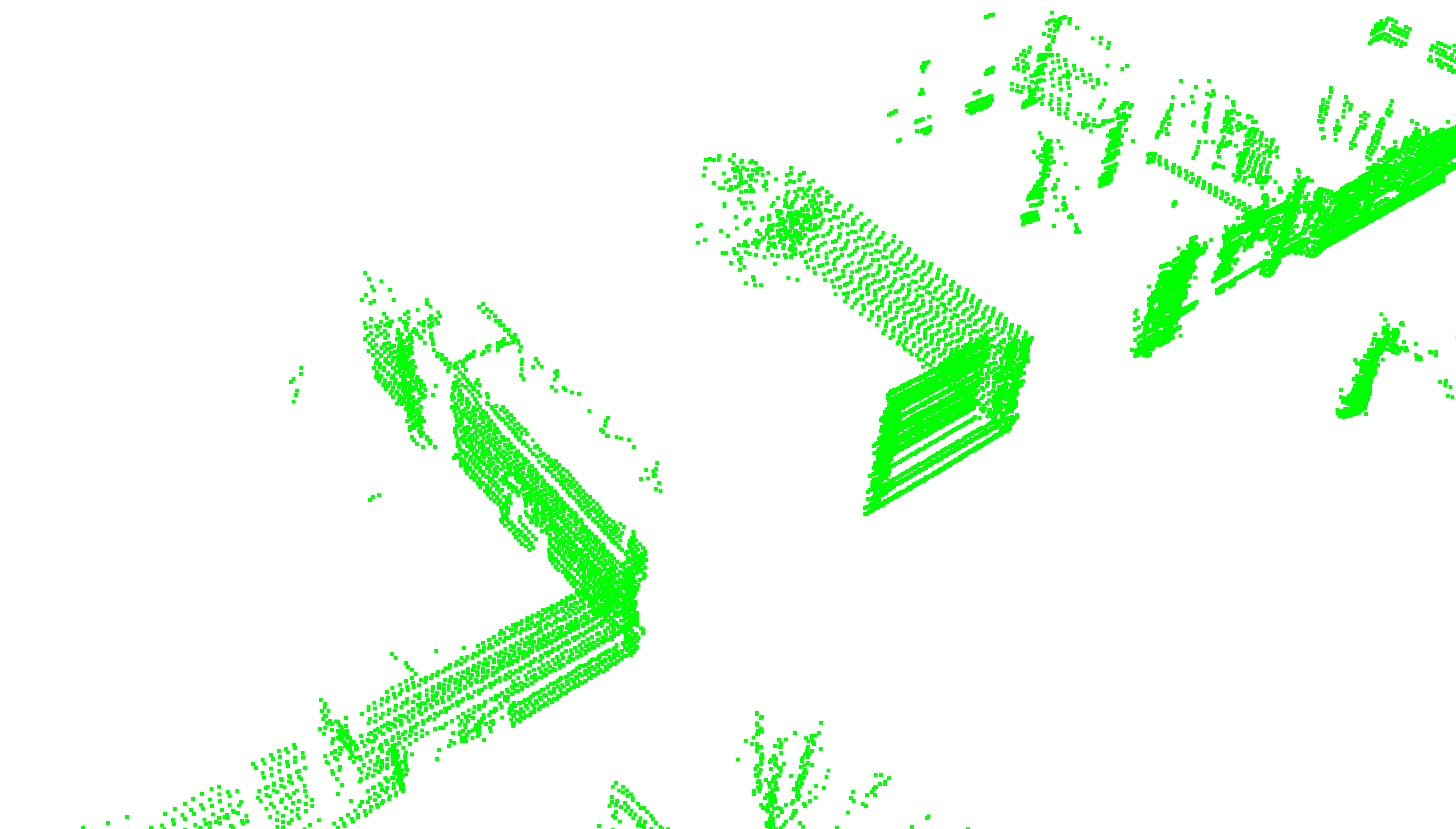}
	}
	\subfloat[Data frame B]{
	  \label{fig:reg_demo_mid} 
	  \includegraphics[width=0.3\linewidth]{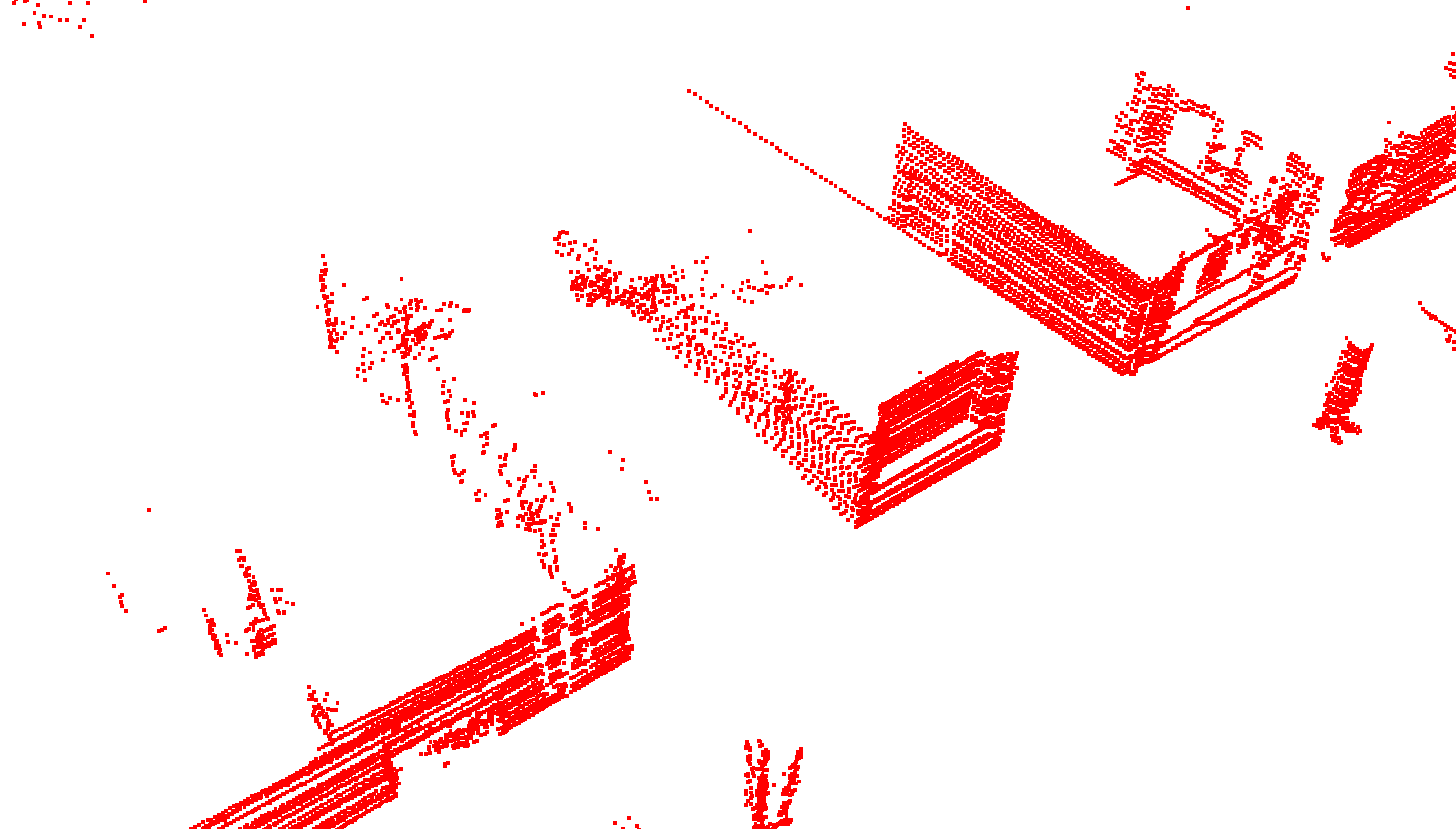}
	}
	\subfloat[Aligned frame]{
	  \label{fig:reg_demo_right} 
	  \includegraphics[width=0.3\linewidth]{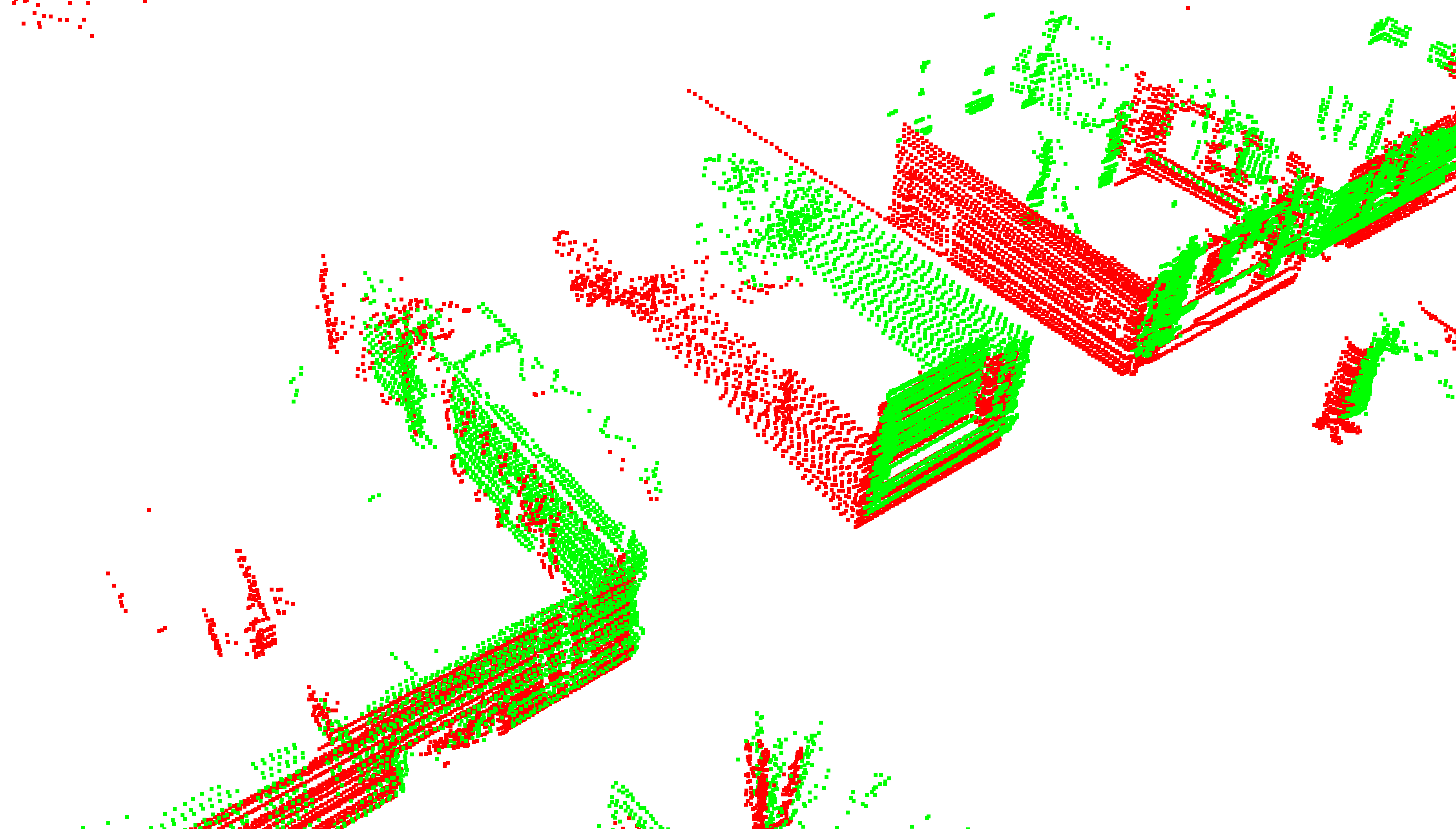}
	}
	\vspace{-5pt}
	\caption{Illustration of point cloud registration. Two point cloud frames are aligned to form a unified frame.}
	\label{fig:reg_demo}
	\vspace{-5pt}
\end{figure}

The single most important building block of 3D perception-enabled applications is \textit{registration}, the process of aligning two frames of point cloud data to form a globally consistent view of the scene.~\Fig{fig:reg_demo} illustrates the registration of two point cloud frames. Augmented Reality applications align a sequence of frames to form a complete 3D model of the environment so as to place virtual objects. Similarly, a mobile robot estimates its real-time position and orientation (a.k.a., odometry) by aligning two consecutive frames, which provides the translational and rotational transformations. As these applications are increasingly deployed in embedded systems with limited performance and power budgets, this paper takes a first step toward enabling real-time, low-power 3D data registration.

We present \proj, a software-hardware system specialized for 3D point cloud registration. \proj achieves high efficiency not only by the specialized datapaths and control logics that mitigate common inefficiencies in general-purpose processors, but also by a combination of acceleration techniques that exploit unique characteristics of point cloud registration. In particular, \proj identifies and exploits different forms of parallelism, captures unique data reuse patterns while reducing the overall compute demand. Critically, we enable these techniques by co-designing the data structure, algorithm, and the accelerator architecture.


We start by understanding the performance characteristics of point cloud registration and identifying the acceleration opportunities. The central challenge, however, is that point cloud registration exposes a large design space with many parameters that are often collectively co-optimized given a particular design target. In order to obtain general conclusions without overly specializing for one particular design point, we first construct a configurable registration pipeline, which let us perform a thorough design space exploration. Surprisingly, although different design points differ significantly in registration accuracy and compute-efficiency, KD-tree search is the single most dominant kernel across all design points, constituting over 50\% of the registration time, and thus presents itself as a lucrative specialization target.

KD-tree search, however, is inherently sequential due to the recursive tree traversal. To enable effective hardware acceleration, we propose a parallel KD-tree search algorithm to introduce fine-grained parallelism that are amenable to hardware acceleration. The algorithm builds on top of the \textit{two-stage KD-tree} data structure, a variant of KD-tree that provides high degrees of parallelism by balancing recursive search with brute-force search. However, two-stage KD-tree necessarily introduces lots of redundant computations in increasing parallelism. To mitigate the redundancies, we observe that point cloud registration is resilient to imprecisions introduced in KD-tree search due to the noisy nature of point cloud data. Our algorithm incorporates an approximate KD-tree search procedure that reduces workload while presenting massive parallelism to the hardware.

The new data structure and algorithm in conjunction uniquely expose two forms of parallelism in KD-tree search: query-level parallelism (QLP) and node-level parallelism (NLP). The key design principle of the hardware accelerator is to exploit the two forms of parallelism with proper architectural mechanisms. Specifically, the accelerator incorporates parallel processing elements (PE) to exploit the QLP while applying pipelining to exploit the NLP within a query. While parallel PEs and pipelining are well-established techniques, effectively applying them in KD-tree search requires us to design a set of architectural optimizations that leverage compute and data access patterns specific to KD-tree search.

We evaluate \proj over a general-purpose system consisting of an Intel Xeon Silver 4110 CPU and an Nvidia RTX 2080 Ti GPU. We show that \proj achieves 77.2$\times$ speedup and 7.4$\times$ power reduction in KD-tree search compared to the GPU, which translates to 41.7\% speedup and 3.0$\times$ power reduction for the end-to-end registration.

To our best knowledge, this is the first paper focusing on architecture and system specializations for point cloud processing. In summary, we make the following contributions:
\begin{itemize}[topsep=2pt]
  \item We identify that KD-tree search is inherently a performance bottleneck in point cloud registration by carefully navigating the algorithmic and parametric design space of registration.
  \item We demonstrate that point cloud registration is tolerant to errors introduced in KD-tree search.
  \item We propose an acceleration-amenable KD-tree search algorithm. Building on top of a novel two-stage KD-tree data structure, the algorithm exposes massive parallelism to the hardware while reducing the compute. 
  \item We co-design an accelerator architecture with the search algorithm. The accelerator incorporates a set of architectural optimizations that are specific to KD-tree search to effectively exploit different forms of parallelism.
\end{itemize}

The rest of the paper is organized as follows.~\Sect{sec:bck} introduces the necessary background of point cloud processing.~\Sect{sec:char} performs extensive algorithmic design space exploration to identify that KD-tree search is the performance bottleneck of point cloud registration.~\Sect{sec:algo} presents an acceleration-amenable KD-tree data structure and search algorithm, and~\Sect{sec:arch} describes the corresponding \proj accelerator architecture.~\Sect{sec:eval} presents the experimental methodology and evaluation results.~\Sect{sec:related} puts \proj in the broad context of related work, and~\Sect{sec:conc} concludes the paper.

%% file: background.tex
\section{Background}
\label{sec:bck}

This section first introduces point cloud data (\Sect{sec:bck:ds}). We then describe point cloud registration, a key task in many application domains that operate on point cloud data (\Sect{sec:bck:reg}).

\subsection{Point Cloud Data}
\label{sec:bck:ds}

Point cloud is a collection of points in a given 3D Cartesian coordinate system. Each point in the point cloud represents the $<x, y, z>$ coordinates of a particular point in the 3D space. Point cloud directly preserves the 3D geometric information of a scene and the spatial relationship between objects of interest, avoiding the need to estimate such information from 2D images. The proliferation of 3D sensors and the emerging interests in 3D geometry-based applications such as robotics lead to massively increased use of 3D data, of which point cloud is the de-facto representation~\cite{aldoma2012tutorial}.

Point cloud data is obtained through 3D sensors, ranging from conventional stereo~\cite{lucas1981iterative} and structured-light cameras~\cite{rocchini2001low} that estimate the scene 3D geometry through computational methods to active sensors such as LiDAR~\cite{schwarz2010lidar} that operate on the "time-of-flight" principles~\cite{gokturk2004time}. While using different mechanisms, different sensors eventually produce the same point cloud data structure. Our paper focuses on the fundamental point cloud processing algorithms, and is independent of how the point cloud data is obtained.





\subsection{Point Cloud Registration}
\label{sec:bck:reg}

A key building block in virtually all point cloud-based applications is \textit{registration}, a process that finds the 4$\times$4 transformation matrix that aligns two point cloud frames to form a globally consistent point cloud. More specifically, given a source point cloud frame $\mathbf{S}$ and a target point cloud frame $\mathbf{T}$, the goal of registration is to estimate a transformation matrix $\mathbf{M}$, which transforms $\mathbf{S}$ to $\mathbf{S'}$ in a way that minimizes the Euclidean distance (i.e., error) between $\mathbf{S'}$ and $\mathbf{T}$. $\mathbf{S'}$ is transformed from $\mathbf{S}$ by applying the transformation matrix $\mathbf{M}$ to every point $X$ in $\mathbf{S}$ to a point $X'$ in $\mathbf{S'}$:


\begin{equation} \label{reg_equ}
	X^{'}_{4\times 1} = \mathbf{M}X_{4\times 1}
	=
	\begin{bmatrix}
	R_{3\times 3}&T_{3\times 1}\\
	0_{1\times 3}&1
	\end{bmatrix}_{4\times 4}X_{4\times 1}
\end{equation}

\noindent where $X = [x, y, z, 1]^{T}$ and $X^{'} = [x^{'}, y^{'}, z^{'}, 1]^{T}$ are the homogeneous coordinates of $X$ and $X'$, respectively. The 4$\times$4 transformation matrix $\mathbf{M}$ consists of a 3$\times$3 rotation matrix $\mathbf{R}$ and a 3$\times$1 translation matrix $\mathbf{T}$, representing all six degrees of freedom.



\paragraph{Significance of Registration} Point cloud registration is a key primitive that finds itself in many application domains.

In many cases point cloud registration \textit{is} the end-to-end machine perception applications such as odometry and mapping. For instance, an autonomous navigation system could capture two consecutive frames $\mathbf{F}_{t}$ and $\mathbf{F}_{t+1}$ in time, and by registering $\mathbf{F}_{t+1}$ against $\mathbf{F}_{t}$ and obtaining the transformation matrix, the navigation system could estimate its own trajectory (rotation and translation) over time, a process known as odometry or ego-motion estimation~\cite{irani1997recovery, zhang2015visual}. Similarly, registration is key to 3D reconstruction~\cite{vosselman20013d}, where a set of frames are aligned against one another and merged together to form a global point cloud of the scene. In other cases point cloud registration is part of the application pipeline to collaborate with other modalities such as camera (e.g., SLAM uses both visual data and point cloud)~\cite{zhang2015visual, park2017colored}.

This paper focuses on improving the efficiency of the core registration operation while being independent of how the high-level applications make use of the registration results.

\begin{figure*}[t]
\centering
\includegraphics[width=2.1\columnwidth]{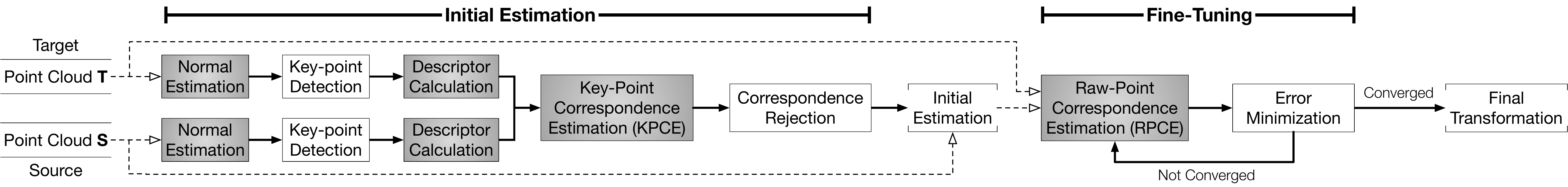}
\caption{The general point cloud registration pipeline, which consists of an initial estimation phase and a fine-tuning phase. The pipeline exposes two kinds of design knobs for accuracy-performance trade-off analysis: algorithmic and parametric choices. Shaded stages make heavy use of KD-tree search, the single-most dominant kernel in all design points.}
\label{fig:pc_pipeline}
\end{figure*}

%% file: char.tex
\section{Performance Characterizations}
\label{sec:char}

\begin{table*}[t]
\Huge
\centering
\caption{Algorithmic and parametric choices and of a general point cloud registration pipeline.}
\label{tab:knobs}
\vspace{-10pt}
\renewcommand*{\arraystretch}{1.1}
\renewcommand*{\tabcolsep}{6pt}
\resizebox{2\columnwidth}{!}
{
\begin{tabular}{cccccc|cc}
\toprule[0.15em]
\multirow{2}{*}{} & \multicolumn{5}{c}{\textbf{Initial Estimation}} & \multicolumn{2}{c}{\textbf{Fine-Tuning}} \\

\textbf{Stages} & \specialcell{Normal\\Estimation} & \specialcell{Key-point\\Detection} & \specialcell{Descriptor\\Calculation} & \specialcell{KPCE} & Rejection & \specialcell{RPCE} & \specialcell{Transformation\\Estimation} \\
 
\midrule[0.05em]
\textbf{\specialcell{Algorithm\\ Choices}} & \specialcell{PlaneSVD\cite{klasing2009comparison}\\AreaWeighted\cite{klasing2009comparison}\\DNN\cite{wang2015designing}} & \specialcell{SIFT\cite{lowe2004distinctive, scovanner20073}\\NARF\cite{steder2011point}\\HARRIS\cite{harris1988combined, sipiran2011harris}} & \specialcell{FPFH\cite{rusu2009fast}\\SHOT\cite{tombari2010unique}\\3DSC\cite{frome2004recognizing}} & \specialcell{-} & \specialcell{Thresholding\\RANSAC\cite{fischler1981random}} & \specialcell{Normal-shooting\\Projection\cite{blais1995registering}} & \specialcell{Error metric\cite{rusinkiewicz2001efficient}\\Solver\cite{rusinkiewicz2001efficient}} \\

\midrule[0.05em]
\textbf{\specialcell{Key\\Parameters}} & Search radius & \specialcell{Scale \\ Range} & Search radius & Reciprocity & \specialcell{Distance threshold\\Ratio threshold} & \specialcell{\# of neighbors\\Reciprocity} & \specialcell{Convergence\\criteria} \\
\bottomrule[0.15em]
\end{tabular}
}
\end{table*}

This section characterizes the performance of point cloud registration through a configurable registration pipeline design that exposes a large accuracy-performance trade-off space (\Sect{sec:char:pipe}). Through an exhaustive exploration of the design space covering both algorithmic and parametric choices, we find that different design points share the same performance bottleneck of KD-tree search, which is thus an ideal acceleration candidate (\Sect{sec:char:dse}). We make the pipeline implementation publicly available at \url{https://github.com/horizon-research/pointcloud-pipeline}.


\subsection{Point Cloud Registration Pipeline}
\label{sec:char:pipe}

Existing implementations of point cloud registration make different trade-offs between accuracy and performance. Intuitively, achieving a higher registration accuracy increases the workload, and vice versa. Our goal in this paper, however, is \textit{not} to overly specialize for  one particular implementation. Rather, we hope to derive general-purpose solutions that benefit different design points.

In order to obtain generally applicable conclusions, a key observation is that different registration implementations, while making different design decisions, all share a similar pipeline substrate. This allows us to construct a general-purpose pipeline with configurable knobs that cover different implementation instances. Critically, our pipeline exposes two kinds of \textit{design knobs} for tuning: algorithmic choices and parametric choices within a particular algorithm.



At the high-level, our pipeline adopts a common two-phase design consisting of an initial estimation phase and a fine-tuning phase~\cite{zhang2014loam, huang2018coarse}. The first phase performs an initial estimation of the transformation matrix, which is then fine-tuned in the second phase until the accuracy converges. The rationale behind the two-phase design is that the fine-tuning phase usually uses an iterative solver to minimize the global registration error; the solver could easily be trapped at local minima if poorly initialized. A carefully designed initial estimation phase would thus significantly improve the efficiency and accuracy of fine-tuning.~\Fig{fig:pc_pipeline} illustrates the high-level architecture of the pipeline, and~\Tbl{tab:knobs} shows the different algorithmic and parametric knobs exposed by the pipeline.


The goal of the initial estimation phase is to calculate an initial transformation matrix by matching a set of salient points from the source point cloud to a set of salient points in the target cloud, similar to image registration~\cite{zitova2003image}, but in the 3D space. 


\begin{figure}[t]
\centering
\subfloat[Translational error vs. time.]
{
\includegraphics[width=0.47\linewidth]{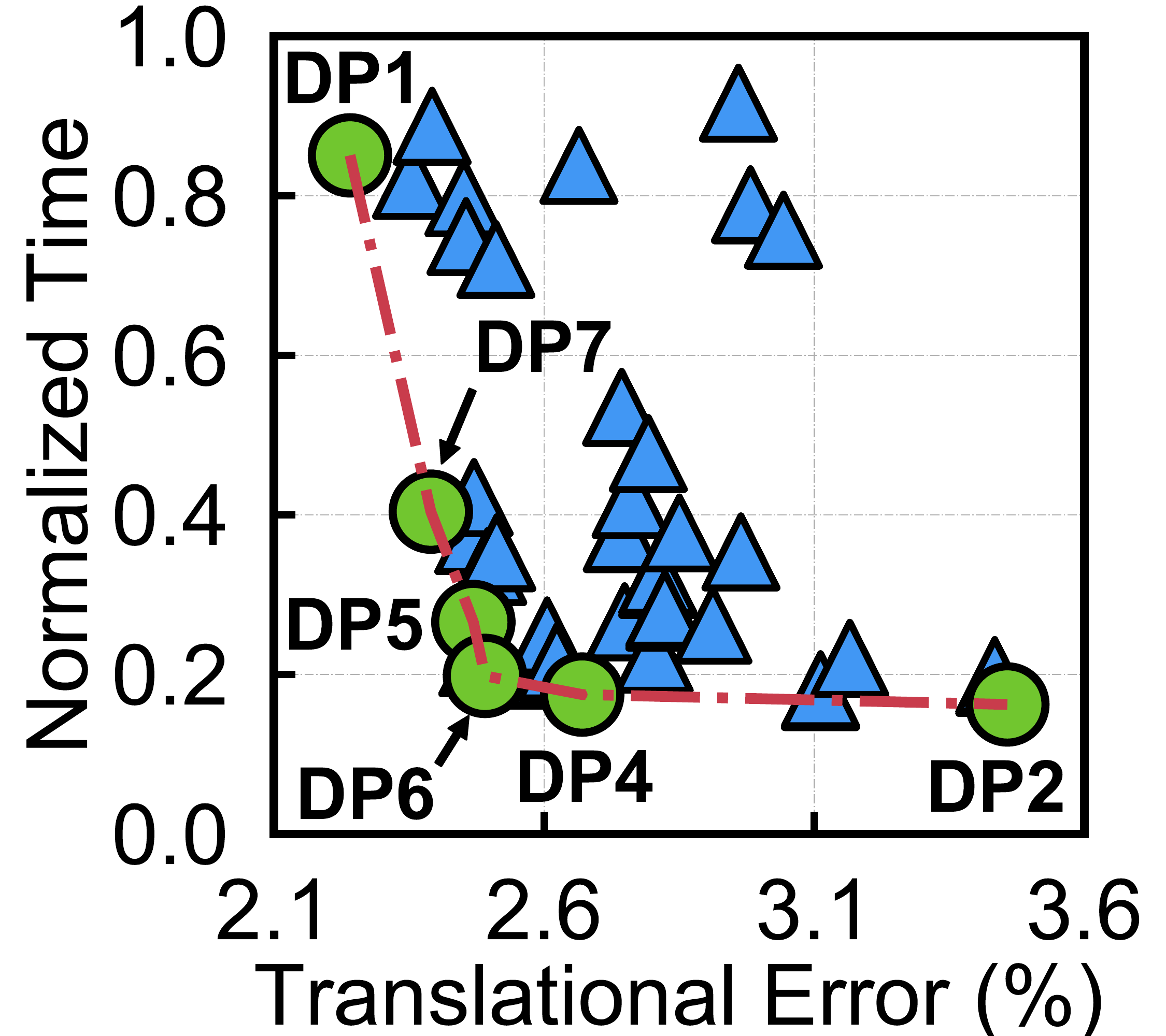}
  \label{fig:dse_tradeoff_te}
}\hfill
\subfloat[Rotational error vs. time.]
{
  \includegraphics[width=0.47\linewidth]{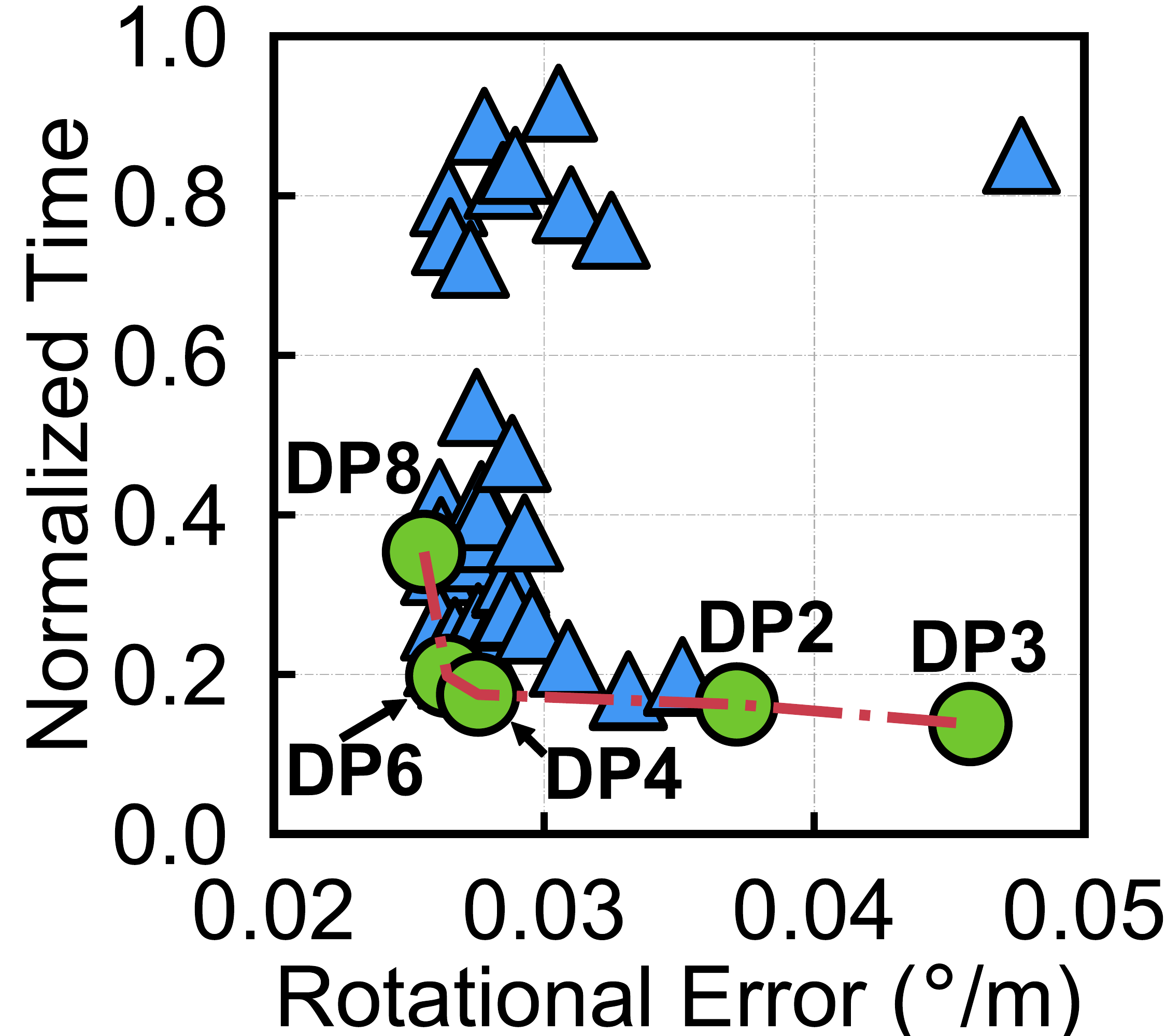}
  \label{fig:dse_tradeoff_re}
}
\vspace{-5pt}
\caption{Quantifying the accuracy-performance tradeoff. We annotate Pareto-optimal design points in both design spaces. Execution time is normalized to \SI{1500}{\milli\second}.}
\label{fig:dse}
\end{figure}

\begin{enumerate}[topsep=2pt]
\item \paragraph{Normal Estimation} The front-end first calculates the surface normal of all points. A point's normal is a 3D vector perpendicular to the tangent plane at the point. Normals are important metadata that will be used in later stages to calculate feature descriptors and to estimate correspondences.

\item \paragraph{Key-Point Detection} This stage selects key-points which contains representative information, from both the target and source point clouds. Operating on the key-points rather than all the points improves the compute-efficiency of the front-end. We explore different feature extraction algorithms such as NARF~\cite{steder2011point} and SIFT~\cite{lowe2004distinctive, scovanner20073}, as well as feature-specific parameters such as the scale of SIFT feature and the range of NARF feature.

\item \paragraph{Feature Descriptor Calculation} This stage computes the feature descriptor of each key-point. A point's feature descriptor is a high-dimensional representation that encodes neighborhood information of the point and therefore provides richer information for registration. Essentially, this stage converts the original 3D point space to a high-dimensional feature space. The dimension of the feature space depends on the specific feature descriptor being used. We explore different descriptors, including FPFH~\cite{rusu2009fast} and SHOT~\cite{tombari2010unique}, as well as key algorithmic parameters such as the search radius when calculating the descriptors.


\item \paragraph{Key-Point Correspondence Estimation (KPCE)} This stage establishes correspondences between the key-points in the source and the target point cloud frames using feature descriptors. Specifically, KPCE establishes the correspondence between a point $\mathbf{s}$ in the source frame and a point $\mathbf{t}$ in the target frame if $\mathbf{t}$'s feature is the nearest neighbor of $\mathbf{s}$' feature in the feature space generated in the previous stage. We explore whether or not reciprocal search is performed.

\item \paragraph{Correspondence Rejection} The final stage of the front-end removes incorrect correspondences produced by the previous stage, and generates a set of correct key-point correspondences, from which the initial transformation matrix $\mathbf{M}$ is estimated. We explore different correspondence rejection algorithms include the classic RANSAC algorithm~\cite{fischler1981random} and ones that simply threshold the distance.

\end{enumerate}

The initial transformation matrix $\mathbf{M}$ allows all points in the source point cloud $\mathbf{S}$ to be transformed to form a new point cloud $\mathbf{S'}$. The fine-tuning phase then estimates the transformation matrix between $\mathbf{S'}$ and the target point cloud $\mathbf{T}$, effectively refining the initial result. The fine-tuning phase uses the popular Iterative Closest Point~\cite{besl1992method, chen1992object} framework, iterating between two stages:

\begin{enumerate}[topsep=2pt]
\item \paragraph{Raw-Point Correspondence Estimation (RPCE)} This stage establishes correspondences between \textit{all} points from the source point cloud $\mathbf{S'}$ and the target point cloud $\mathbf{T}$. For every point in $\mathbf{S'}$, RPCE finds its nearest neighbor in $\mathbf{T}$. Different from KPCE, RPCE searches in the original 3D point space.
\item \paragraph{Transformation Estimation} This stage formulates an error measure between every pair of corresponding points identified previously, and minimizes the error using an optimization solver, which produces the transformation matrix $\mathbf{M'}$ between $\mathbf{S'}$ and $\mathbf{T}$, and transforms $\mathbf{S'}$ into $\mathbf{S''}$. $\mathbf{S''}$ then becomes the new source point cloud, and is fed back to the RPCE stage. We explore different error formulations (e.g., mean square point-to-point~\cite{johnson1999registration} or point-to-plane error~\cite{chen1992object}) and different solvers including the Singular Value Decomposition~\cite{golub1971singular} and the Levenberg-Marquardt algorithm~\cite{more1978levenberg}. Another key parameter that we explore is the convergence criteria, which determines the termination of ICP, and thus impacts both the accuracy and compute time.
\end{enumerate}

\begin{figure}[t]
\centering
\subfloat[Distribution across the seven key stages (\Fig{fig:pc_pipeline}).]
{
  \includegraphics[trim=0 0 0 0, clip, width=.9\columnwidth]{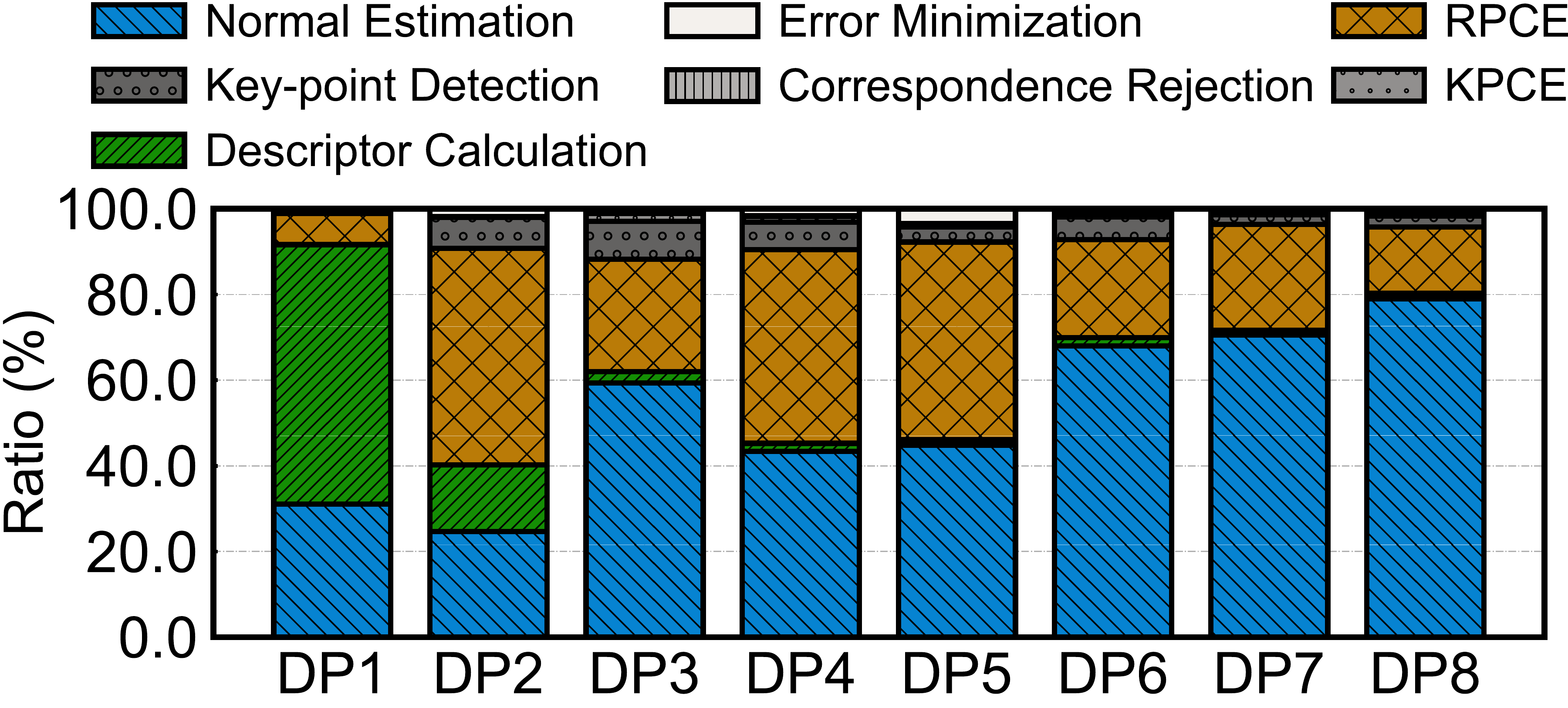}
  \label{fig:stage_dist}
}\\
\subfloat[Distribution between KD-tree search and other operations.]
{
  \includegraphics[trim=0 0 0 0, clip, width=.9\columnwidth]{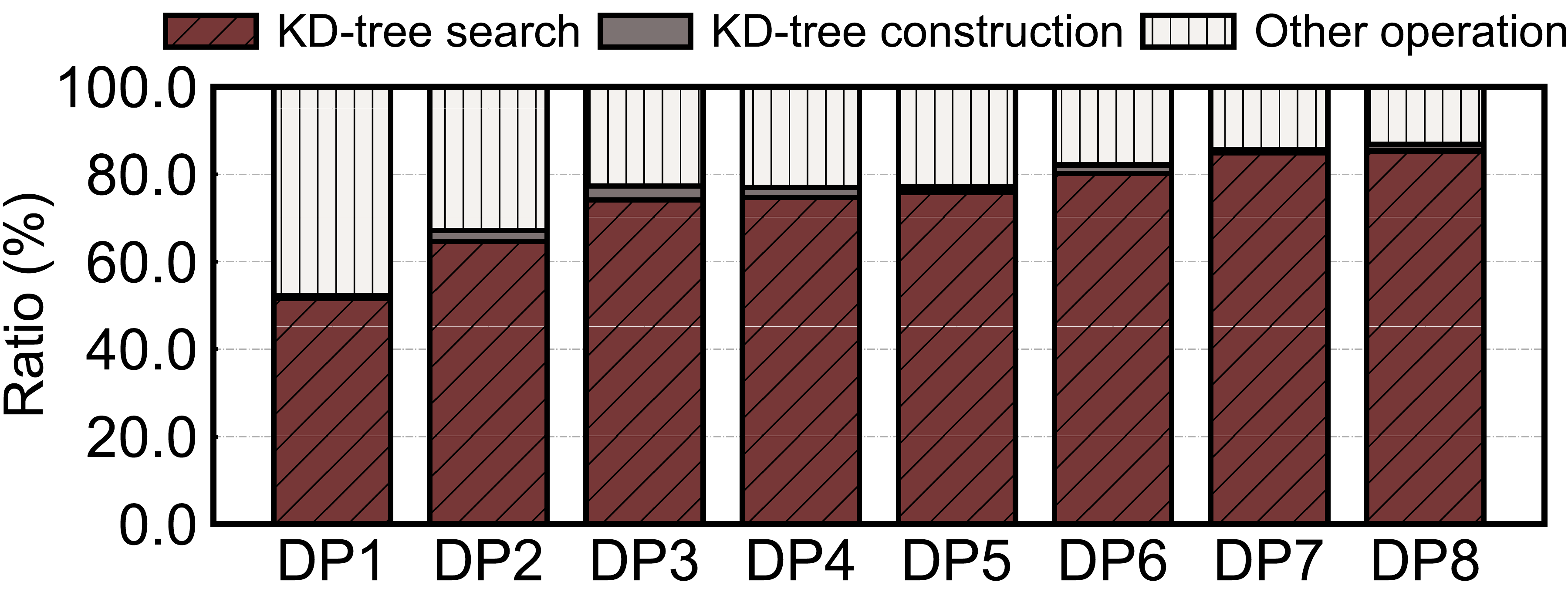}
  \label{fig:kernel_dist}
}
\caption{Time distribution of point cloud registration of the eight Pareto-optimal design points (denoted as DP$i$) obtained from the design spaces in~\Fig{fig:dse_tradeoff_te} and~\Fig{fig:dse_tradeoff_re}.}
\label{fig:dse}
\end{figure}

\subsection{Performance Bottleneck Analysis}
\label{sec:char:dse}

\begin{figure*}[t]
	\centering
    \subfloat[Canonical KD-tree data structure.]{
      \label{fig:kdtree}
      \includegraphics[width=0.47\linewidth]{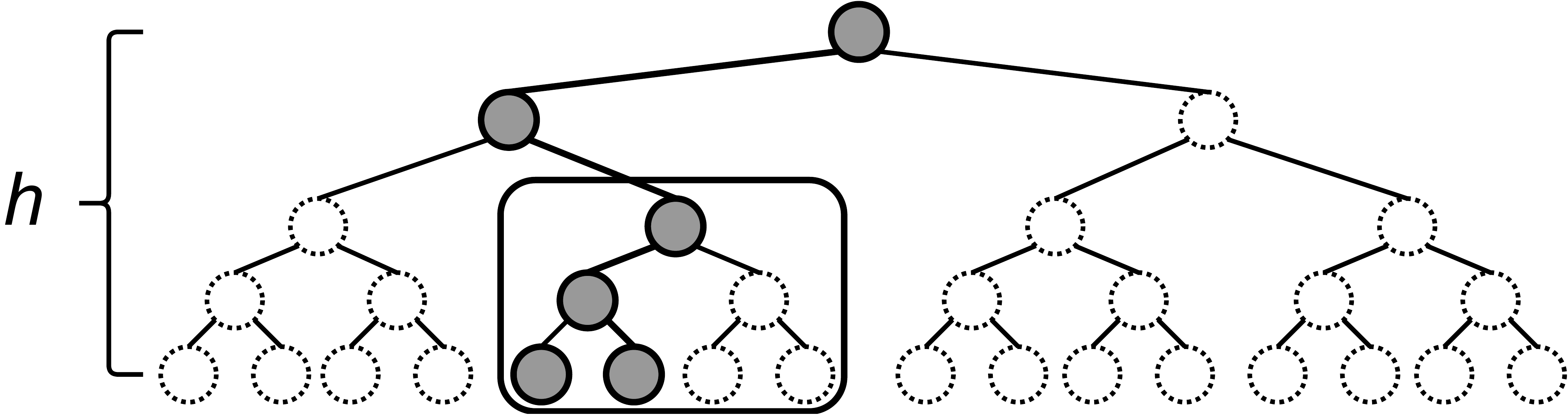}
    }
    \hfill
    \subfloat[Two-stage KD-tree data structure.]{
      \label{fig:2skdtree}
      \includegraphics[width=0.47\linewidth]{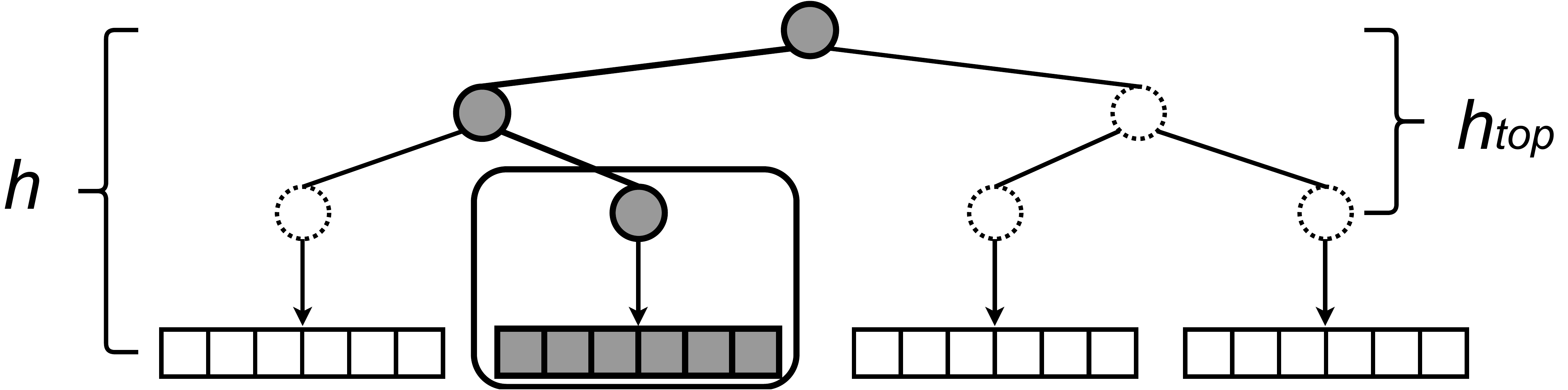}
    }
    \vspace{-5pt}
    \caption{Comparison between the canonical and the two-stage KD-tree data structures. Shaded nodes are visited during the search while the rest of the nodes are pruned. The top-tree in the two-staged data structure is exactly the same as the corresponding portion in the classic data structure. Each leaf node in the top-tree organizes its children as an unordered set rather than a sub-tree to enable exhaustive search. While exposing parallelism, the two-stage data structure requires visiting more nodes: nine nodes as opposed to six nodes required by the classic data structure in this example.}
    \label{fig:ds}
    \vspace{-5pt}
\end{figure*}

\paragraph{Design Space Exploration} Using the configurable pipeline, this section performs a design space exploration (DSE) to identify representative design points, on which we then study the performance bottlenecks. The design space is specified by the different algorithmic choices and parameter values described in~\Tbl{tab:knobs}. We use the widely-adopted KITTI dataset~\cite{geiger2012we} and perform the experiments on a Xeon 4110 processor (see~\Sect{sec:eval:exp} for detailed experimental setup).~\Fig{fig:dse_tradeoff_te} shows how different design points trade translation error for execution time, and~\Fig{fig:dse_tradeoff_re} shows the trade-off between rotational error and execution time. The DSE results confirm the vast trade-offs space exposed by our configurable pipeline. More importantly, we are able to identify the Pareto-optimal frontier in each design space as annotated both in ~\Fig{fig:dse_tradeoff_te} and ~\Fig{fig:dse_tradeoff_re}. To draw meaningful conclusions, we now focus on analyzing the Pareto-optimal design points from both design spaces.




\paragraph{Performance Bottleneck} Our goal is to identify ``universal'' performance bottlenecks that, if accelerated, would lead to speed improvements on a wide range of design points rather than being overly tied to a particular design point.

To that end, we first examine the per-stage performance of the pipeline.~\Fig{fig:stage_dist} shows the registration time distribution across the seven key stages as described in~\Fig{fig:pc_pipeline} for the eight Pareto-optimal design points (DP). Normal Estimation, Descriptor Calculation, and RPCE are three dominating stages, constituting to over 90\% of the total time. However, there is no single dominant stage that is consistent across different design points. For instance, while the Normal Estimation stage contributes to about 80\% of the execution time in DP8 and thus is an ideal acceleration target, it contributes to less than 30\% of the execution time in DP1 and DP2. The diversity of stage-wise time distribution indicates that accelerating any single stage would not yield a general solution.

Looking into the operations within each stage, however, we find that Normal Estimation, Descriptor Calculation, and RPCE all make heavy use of neighbor search. For instance, to calculate the surface normal for a given point in the Normal Estimation stage, one must identify the neighbors of the given point in order to form a surface, with which the normal is calculated. Similarly, the very definition of ``correspondence'' in the RPCE stage requires identifying the nearest neighbor of a given point. In particular, KD-tree is arguably the most efficient data structure that is widely used in neighbor search, providing an average time complexity of $O(\log{}n)$~\cite{bentley1975multidimensional, finkel1977algorithm}. The majority of the point cloud registration implementations use KD-tree for neighbor search~\cite{greenspan2003approximate, li2016tree, Wang2017Point, Shi2016ICPKD}. We thus equate KD-tree search with neighbor search in the rest of the paper.

As a result of the inherently algorithmic requirement of different pipeline stages, KD-tree search is a key operation that dominates the registration time across different DPs.~\Fig{fig:kernel_dist} shows that the KD-tree search operation consistently contributes to 50\% - 85\% of the total time in all the design points. Accelerating KD-tree would thus be a key performance optimization that is generally applicable to different point cloud registration implementations.

%% file: algo.tex
\section{Acceleration-Amenable KD-Tree Data Structure and Algorithm}
\label{sec:algo}

KD-tree search is inherently sequential as it requires tree traversal~\cite{bentley1975multidimensional}. To enable hardware acceleration, we propose a mechanism that exposes massive parallelism in KD-tree search while reducing the total compute -- at the cost of negligible end-to-end accuracy loss. The key is to co-design the KD-tree data structure with a new approximate search algorithm. This section first describes a parallelism-exposing KD-tree data structure (\Sect{sec:algo:ds}). We then quantify the error-tolerating nature of KD-tree search (\Sect{sec:algo:error}), and describe our new search algorithm (\Sect{sec:algo:approx}).


\subsection{Two-Stage KD-Tree Data Structure}
\label{sec:algo:ds}

We first briefly describe the classic KD-tree data structure and its associated search algorithm. We then describe the two-stage KD-tree data structure, which exposes higher degrees of parallelism during search while introducing redundant work.

\paragraph {Canonical KD-Tree} A KD-tree is a data structure that organizes points in a k-dimensional space in a binary search tree  to enable efficient search~\cite{bentley1975multidimensional}. Each tree node stores a k-dimensional point. The point on each non-leaf node implicitly generates a splitting hyperplane that divides the space into two half-spaces. Points that lie in the left half-space are stored in the left sub-tree, and points that lie in the right half-space are stored in the right sub-tree. Essentially, each non-leaf node corresponds to a bounding box in the k-dimensional space that encapsulates all the nodes in its sub-tree. Usually the median point is used to generate the splitting plane such that the resulting KD-tree is a balance tree.


Point cloud registration mainly involves two kinds of search: radius search and Nearest Neighbor (NN) search. Given a query, which itself is also a point in the k-dimensional space, the former returns all the points in the point cloud that are within the given radius to the query point, and the latter returns the nearest neighbor to the query point. Without losing generality, we use NN search to drive the explanation.

The KD-tree search algorithm starts from the root node, and recursively traverses the tree using the query point. As the algorithm visits a node, it checks whether the node should be added to the return results by comparing against the current nearest distance $d$. The algorithm then further searches the left and right sub-tree of the current node. Critically, if the bounding box of either sub-tree does not intersect with the hypersphere surrounding the query point with the $d$, the entire sub-tree could be skipped because all of its nodes are guaranteed to lie outside of $d$. This is a key technique called \textit{pruning} that enables efficient search in KD-tree.~\Fig{fig:kdtree} shows a simple KD-tree example, where the shaded points are visited during the search while the rest of the points are pruned.


While pruning reduces redundant computations by skipping unnecessary nodes, it serializes the search: every time the algorithm visits a node, it might obtain a new current nearest distance, which allows for pruning more nodes later.

\paragraph {Two-Stage KD-Tree} To balance parallelism and redundancies, we use a slight variant of the canonical KD-tree data structure called two-stage KD-tree.~\Fig{fig:2skdtree} shows the two-stage KD-tree organization of the same points stored in the canonical KD-tree (\Fig{fig:kdtree}). The two-stage KD-tree is split into two halves. The top half, which we call the top-tree, is a tree with height $h_{top}$. The top-tree is exactly the same as the first $h_{top}$ levels of the classic KD-tree. Each top-tree leaf node organizes its children as an unordered set as opposed to a sub-tree as in the canonical data structure.




Since the leaf nodes of the top-tree organize their children as unordered sets, different child nodes of a leaf node can be searched in parallel. Essentially, the two-stage KD-tree enables exhaustive searches in certain sub-trees. In the extreme case where $h_{top}$ is 0, searching in the two-stage KD-tree is equivalent to exhaustively searching all the points. Fundamentally, the two-stage KD-tree introduces more parallelism at the cost of higher redundancies compared to the canonical KD-tree data structure. In the example of~\Fig{fig:ds}, searching in the two-stage data structure visits nine nodes, three when traversing the top-tree and six when exhaustively searching a leaf node of the top-tree, as opposed to six nodes required by the classic data structure.




\begin{figure}[t]
	\centering 
    \subfloat[Redundancy ratio.]{
      \label{fig:redundancy_ratio} 
      \includegraphics[width=0.47\linewidth]{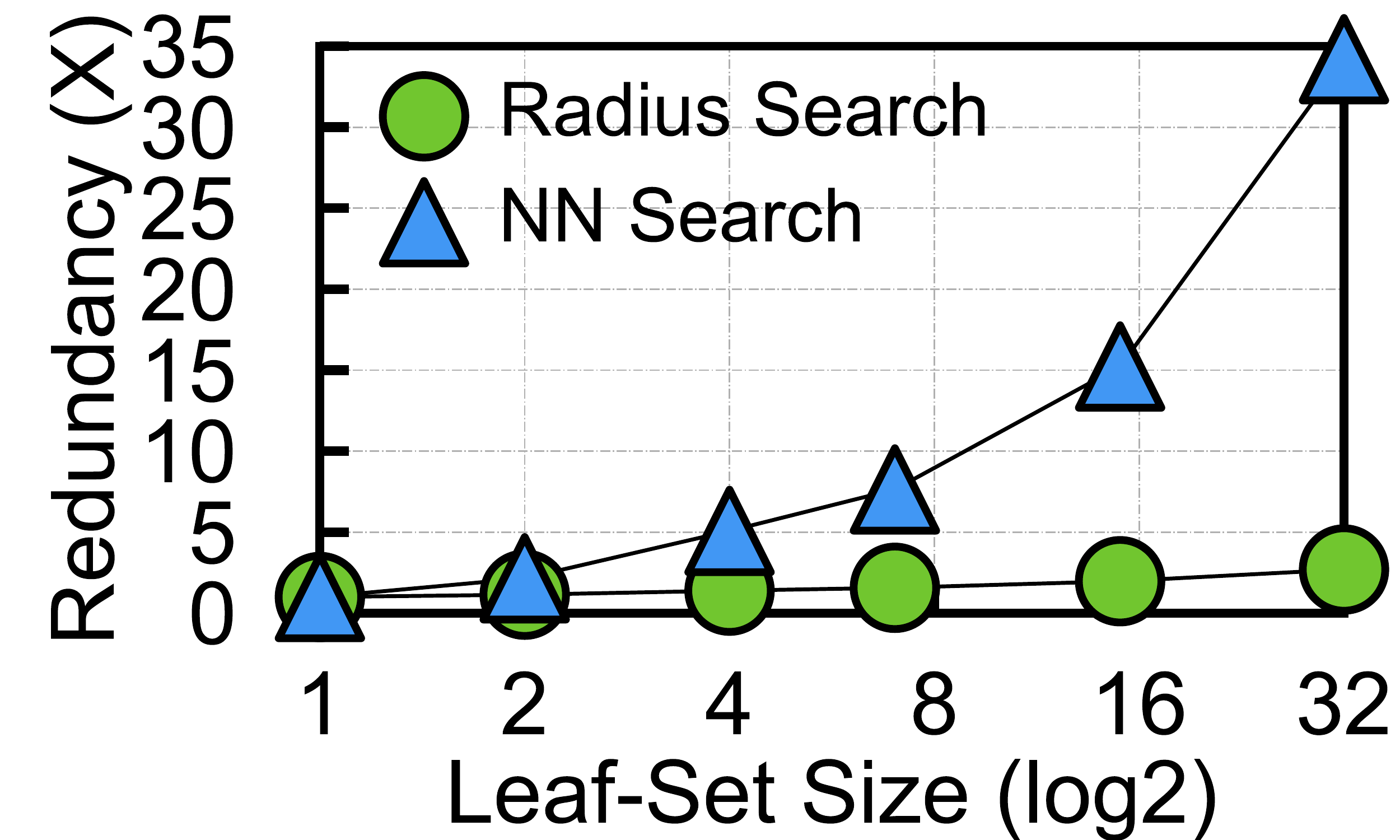}
    }
    \subfloat[Total number of nodes visited.]{
      \label{fig:redundancy_abs}
      \includegraphics[width=0.47\linewidth]{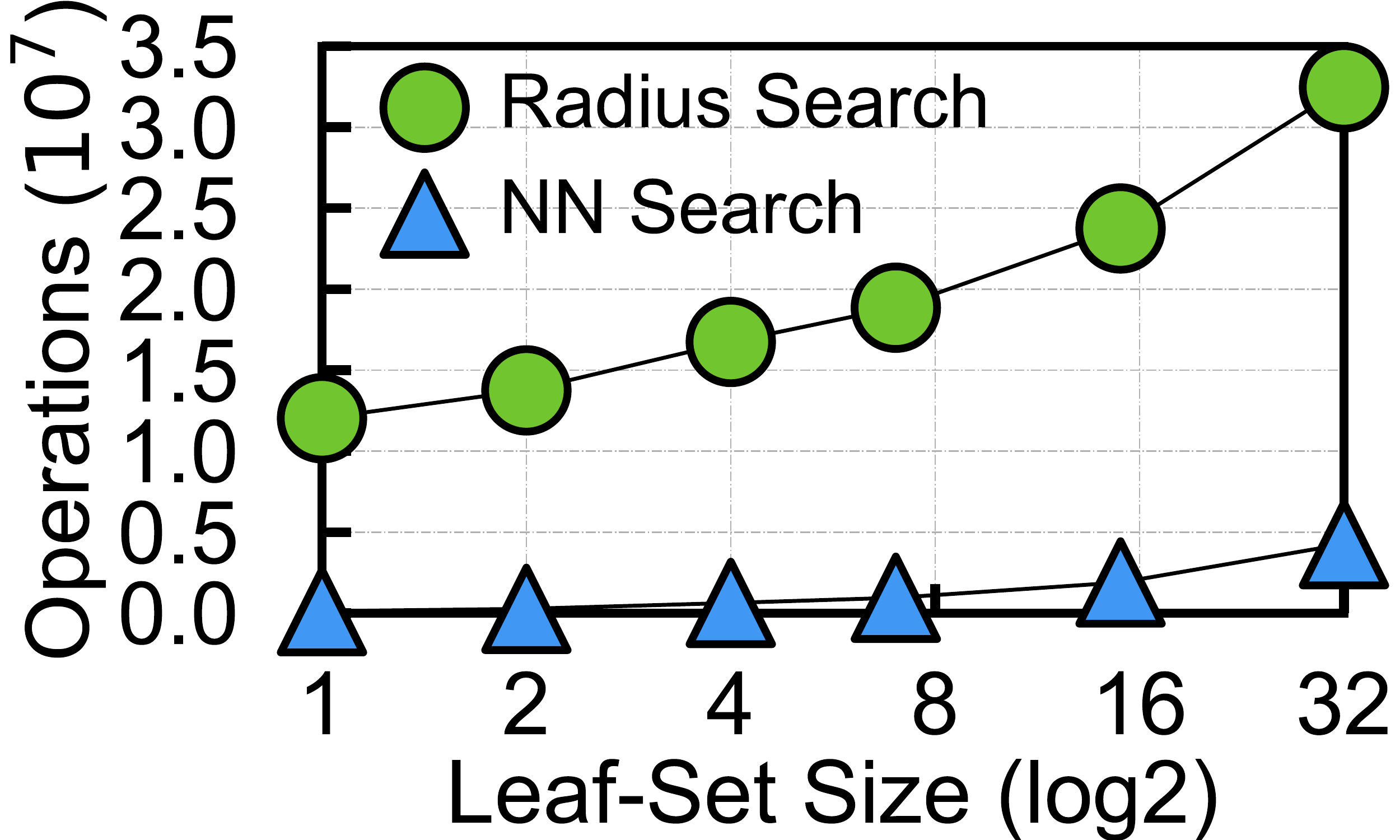}
    }
    \vspace{-5pt}
    \caption{The two-stage KD-tree introduces redundant visits to nodes. Redundancy is quantified as the ratio between the number of nodes visited in the two-stage KD-tree and that in the classic KD-tree. Redundancy increases as the leaf-set size grows. The leaf-set size is defined as the number of children in the leaf node's unordered set.}
    \label{fig:redundancy}
\end{figure}

Intuitively, a shorter top-tree exposes more parallelism but also introduces more redundancies. Using the KITTI Odometry Dataset (see~\Sect{sec:eval:exp} for the detailed experimental setup),~\Fig{fig:redundancy_ratio} shows how the redundancy introduced by the exhaustive searches varies with the leaf-set size for both radius search and NN search. The redundancy is quantified as the ratio between the number of nodes visited in the two-stage KD-tree and that in the classic KD-tree. The leaf-set size is defined as the number of children in the leaf node's unordered set. The classic KD-tree has a leaf-size one, and the two-stage KD-tree in~\Fig{fig:2skdtree} has a leaf-set size six.

As the leaf-set size increases, the top-tree height decreases and more exhaustive searches occur. Thus, the redundancy increases. With a leaf-set size of 32, the two-stage KD-tree introduces about 35$\times$ redundant node visits for NN search and about 3$\times$ for radius search. The redundancy grows much faster for the NN search than for the radius search because the NN search benefits more from pruning than the radius search, and thus suffers more from exhaustive searches. While the redundancy introduced to radius search seems lower than NN search, the sheer number of nodes that radius search has to visit is much greater than NN search as shown in~\Fig{fig:redundancy_abs}, which shows the absolute number of nodes visited as the leaf-set size increases. Thus, the redundancies introduced by the two-stage KD-tree is significant for radius search as well.



\subsection{Quantifying the Error-Tolerance}
\label{sec:algo:error}

\begin{figure}[t]
    \centering
    \captionsetup[subfigure]{width=0.47\columnwidth}
	\subfloat[Sensitivity of translational error as the NN search returns the $k^{th}$ nearest neighbor instead of the nearest neighbor.]
	{
  		\label{fig:sensitivity_k}
  		\includegraphics[width=0.47\linewidth]{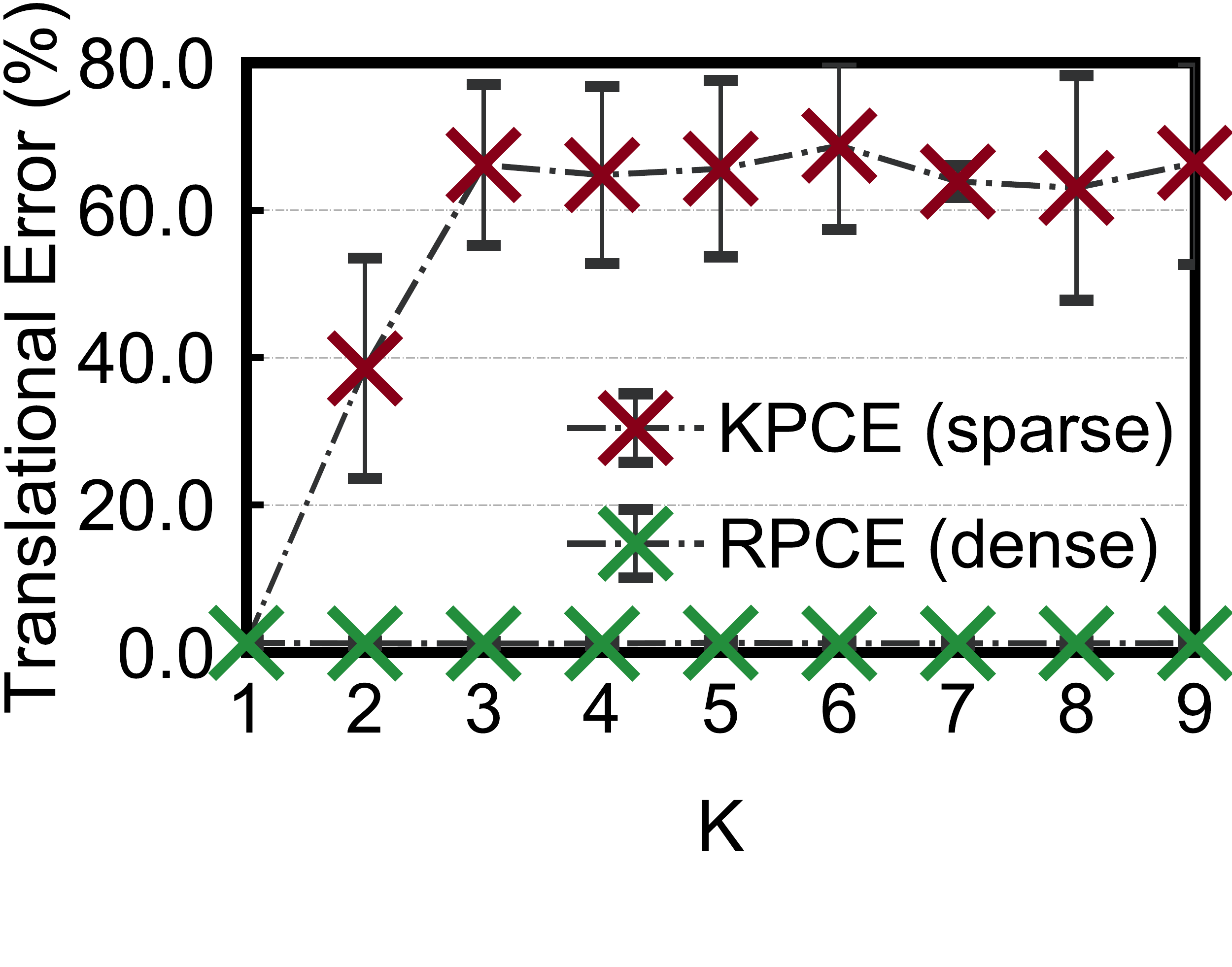}
	}\hfill
	\subfloat[Sensitivity of translational error as the radius search returns points between <$r1, r2$> rather than within $r$ ($r$ = \SI{60}{\centi\meter} here).]
	{
  		\label{fig:sensitivity_ring}
  		\includegraphics[width=0.47\linewidth]{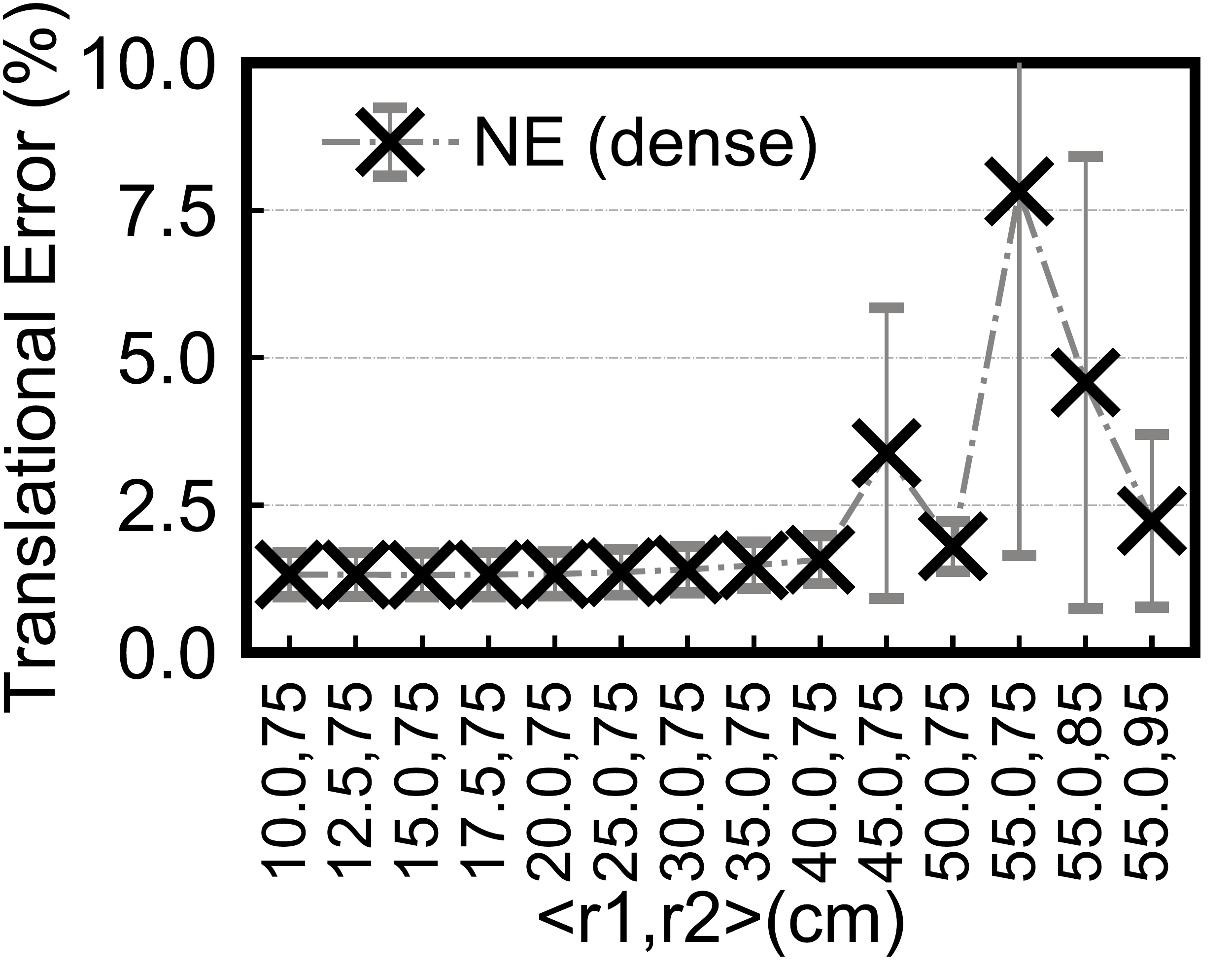}
	}
	\vspace{-5pt}
    \caption{Registration error ($y$-axis) varies as the degree of error ($x$-axis) changes. The error is robust against inexactness of KD-tree search when searching dense points (NE and RPCE), but is sensitive when searching sparse points (KPCE).}
\end{figure}

While the two-stage KD-tree data structure exposes more parallelism, it also introduces lots of redundancies to the search on leaf nodes. To mitigate the redundancies, our key observation is that KD-tree search does not have to be exact because the entire point cloud registration pipeline is error-tolerant. By performing inexact searches on the two-stage KD-tree, we could reduce the amount of computations while retaining parallelism. This section quantitatively demonstrates the error resilience while leaving the mechanisms to exploit the resilience to the next section.

There are two reasons that point cloud registration is resilient to inexact KD-tree search. First, acquiring point cloud data is inherently an approximation process due to the sensor noise. The movement of the sensor during acquisition further adds uncertainties to data acquisition. Second, the registration algorithm strives to minimize the global error, where local inexactness would be compensated at the global scale.


\paragraph{Error Injection} To understand the impact of inexact KD-tree search on the registration accuracy, we manually inject errors into the KD-tree search, and quantify how the end-to-end registration accuracy varies with the KD-tree search accuracy. Specifically, we inject errors into the nearest neighbor (NN) search by replacing the return result, i.e., the nearest neighbor to the query, with a point that is the $k^{th}$ nearest neighbor to the query point. Similarly, we inject errors into radius search by replacing the return results, i.e, points that lie within a sphere delineated by the radius $r$, with points that lie within a spherical shell delineated by two radiuses $r1$ and $r2$, where $r1 < r < r2$. The parameters $k$ and $<r1$, $r2>$ control the degrees of error injected into the radius search and the NN search on KD-tree, respectively.



\SetKw{Continue}{continue}
\begin{algorithm2e}[t]
\SetAlgoLined
\KwIn{QuerySet $\mathbb{Q}$; LeafNode $LF$; Threshold $thd$.}
\KwResult{Search result $q.res$ for all $q$ in $\mathbb{Q}$.}

\For {$q$ in $\mathbb{Q}$} {
 \If {$LF.leaders.size()$} {
   \tcp{Find the closest leader for $q$}
   $closestLeader = getMinDist(q, LF.leaders)$ \\
   \If {$dist(q, closestLeader) < thd$} {
     \tcp{Approximate path: search in the results of $closestLeader$}
     $q.res = bf{\text -}search(q, closestLeader.res)$ \\
     \Continue; \\
   }
 }
 \tcp{Precise path: search in all the children of the leaf node $LF$}
 $q.res = bf{\text -}search(q, LF.children)$ \\
 $LF.leaders.pushback(q)$ \\
}
\caption{Approximate KD-Tree Search.}
\label{alg:approxkd}
\end{algorithm2e}

\paragraph{Error Tolerance} While multiple stages make use of KD-tree search, we mainly inject errors into two stages: Normal Estimation (NE) and Raw-Point Correspondence Estimation (RPCE), both contributing heavily to the total execution time (\Fig{fig:stage_dist}). The former uses radius search and the latter uses NN search.~\Fig{fig:sensitivity_k} and~\Fig{fig:sensitivity_ring} show how the end-to-end registration error varies with different degrees of error injected into RPCE and NE, respectively. Due to space limit, we show only the translational error; the trend on rotational error is similar. Error bars denote the standard deviation of all the frames' errors in one sequence.


We find that the registration error is statistically robust to errors introduced in both the radius search and NN search, indicating the potential of relaxing KD-tree search accuracy. For instance, the registration error is virtually the same if the radius search returns the points between $<30, 75>$ compared to the precise search that returns points within $r=65$.


Critically, not all instances of KD-tree search are equally amenable to approximation. While the NE stage and the RPCE stage both operate on dense points, we find that errors introduced in KD-tree search that operates on sparse data are detrimental to registration accuracy. For instance, the Key-Point Correspondence Estimation (KPCE) stage operates on sparse (feature) data.~\Fig{fig:sensitivity_k} overlays how the registration accuracy varies with the error degree introduced in the KPCE stage. Returning just the second nearest neighbor leads to about 40\% registration accuracy loss.


Overall, we find that KD-tree searches that operate on dense points are amenable to approximation, and thus provide an opportunity to greatly reduce the amount of computations in the KD-tree search. We particular focus on the NE and RPCE stages as they dominate the end-to-end performance.


\subsection{Approximate KD-Tree Search}
\label{sec:algo:approx}

Motivated by the error resilience of the point cloud registration pipeline, we propose an approximate KD-tree search algorithm that reduces computation overheads with little accuracy loss. Our key observation is that queries arriving at the same leaf node in the top-tree are close to each other as they fall into the same 3D partition. Therefore, it is likely that their search results are similar.

Leveraging this insight, our idea is to split queries arriving at the same leaf node into a \textit{leaders} group and a \textit{followers} group. Queries in the leaders group perform an exhaustive search in the leaf node's children as usual, while queries in the followers group search in only the return results of the closest leader. To dynamically adjust the leaders group, we introduce a discriminator $thd$; if the distance between a query point and the closest leader is greater than $thd$, the query point is added to the leaders group.~\Alg{alg:approxkd} shows the pseudo-code of the algorithm.

This algorithm relies on an efficiency trade-off: it allows a follower query to search in a much smaller space, i.e., its closest leader's neighboring points as opposed to all the children of the leaf node, while incurring the cost to find the closest leader. Assuming that one leaf node has $N$ children points; there are $L$ points in the leaders group, and the returned neighbors of a leader consists of $R$ points. A follower query would compare against $L+R$ points, which should be much smaller than $N$ for the algorithm to succeed. This first-order cost model is used to understand the performance gains in \Sect{sec:eval:res}.

%% file: arch.tex
\section{KD-Tree Accelerator Design}
\label{sec:arch}

\begin{figure}[t]
\centering
\includegraphics[width=1\columnwidth]{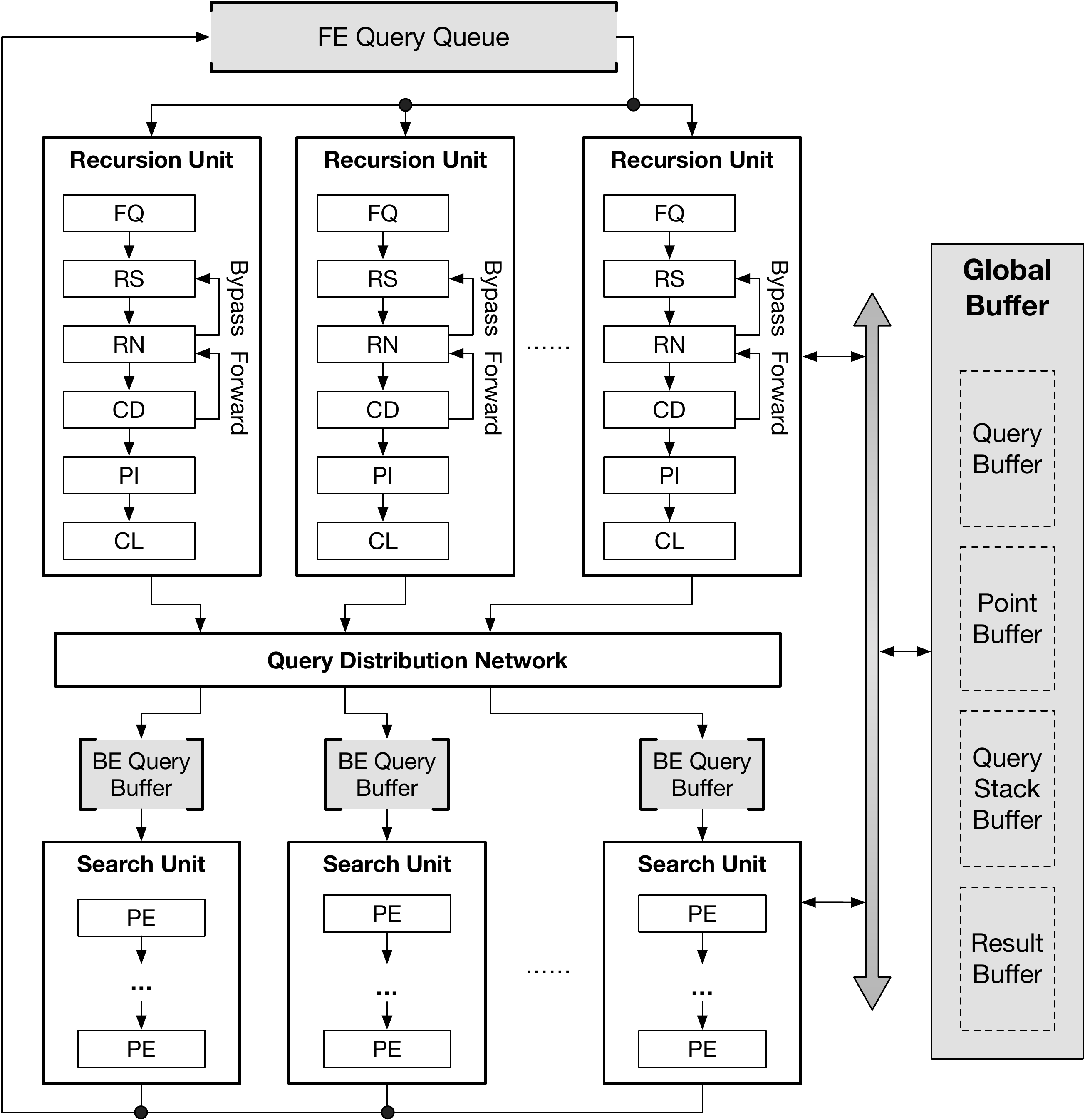}
\caption{The \proj accelerator architecture overview. The front-end (FE) consists of a set of recursion units (RU) that process queries in the top-tree. The back-end (BE) consists of a set of search units (SU) that process queries by exhaustively searching children in the leaf nodes. Queries are distributed from the FE to the BE buffers, and the BE reinserts queries to the FE query queue. The global buffer maintains all the necessary metadata.}
\label{fig:arch}
\end{figure}

This section describes the accelerator design. We first provide an overview of the architecture (\Sect{sec:arch:ov}). We then describe its two key components: the front-end (\Sect{sec:arch:ru}) and the back-end (\Sect{sec:arch:su}).

\begin{figure*}[t]
\centering
\includegraphics[width=2\columnwidth]{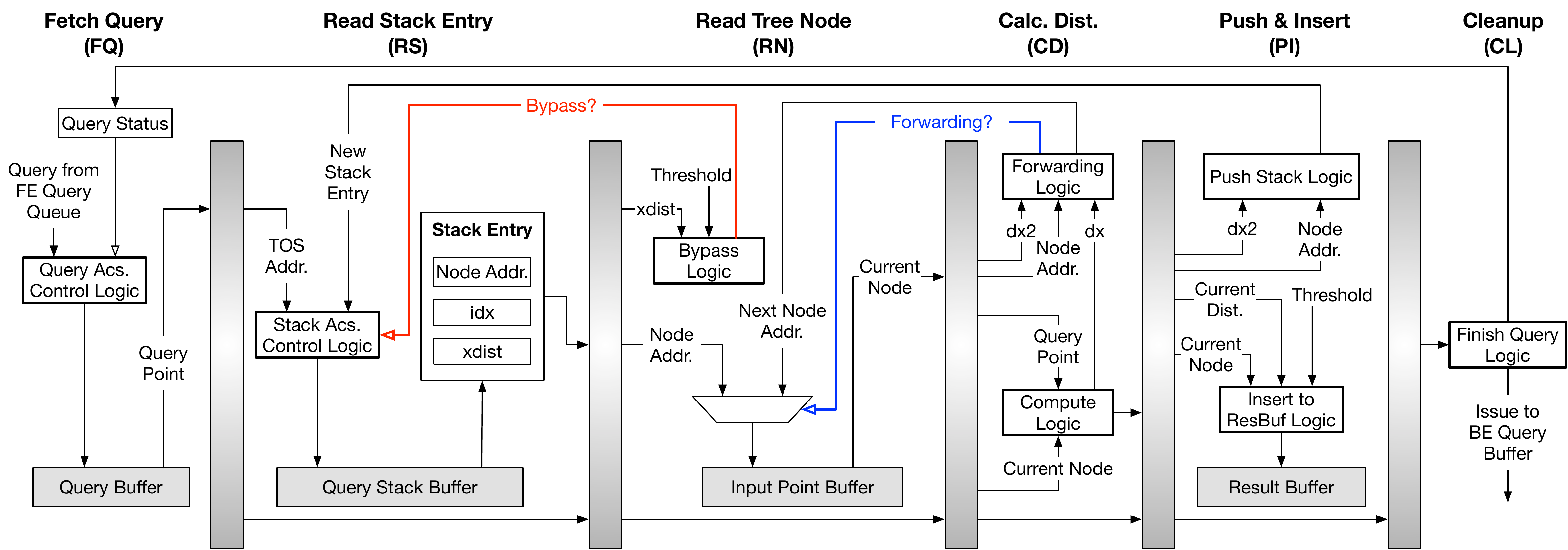}
\caption{Recursion unit (RU) pipeline overview. Each RU recursively traverses the top-tree in a DFS manner for one query at a time. The traversal status is maintained by a stack, introducing data dependencies between the \textbf{PI} and the \textbf{RS} stages. Node bypassing and forwarding eliminate the pipeline stalls.}
\label{fig:ru}
\end{figure*}

\subsection{Accelerator Overview}
\label{sec:arch:ov}

The new data structure and search algorithm expose two levels of parallelism. First, each query can be processed in parallel, both in searching the top-tree and in exhaustively searching leaf nodes. We call it query-level parallelism (QLP). Second, in the exhaustive search stage, different child nodes could be processed in parallel within each query. We call it node-level parallelism (NLP). The hardware architecture is designed to provide the mechanisms to support the two forms of parallelism while exploiting the data locality.~\Fig{fig:arch} shows an overview of the accelerator.

The accelerator consists of a Front-End (FE)  that is responsible for searching in the top-tree and a Back-End (BE) that is responsible for searching in the leaf nodes. The FE uses a set of Recursion Units (RU), each processing one query at a time, to exploit QLP in the top-tree. Each incoming query is first inserted into the FE Query Queue (FQQ), from which each RU deques a query to search in the top-tree. Once the top-tree search for a query is finished, the RU sends the query to the BE. The BE uses a set of Search Units (SU), each responsible for a set of leaf nodes. Queries coming from the FE are first inserted into an SU's BE Query Buffer (BQB), from which the SU schedules the queries to execute. Each SU has a set of Processing Elements (PEs), exploiting both QLP and NLP.

Processing search queries requires a set of input and output metadata, which is stored in a global buffer. Specifically, the buffer is partitioned to hold the following metadata: (1) an Input Point Buffer that holds all the points in the point cloud, (2) a Query Buffer that holds all the query points, (3) a Result Buffer that holds the return results, and (4) a Query Stack Buffer that holds the recursion stacks for all the queries. We reserve the maximal number of stack entries for each query (i.e., the height of the top tree) in the buffer.


\subsection{The Front-End: Recursion Unit}
\label{sec:arch:ru}

The FE processes queries in the top-tree. For each query, the FE recursively searches the top tree until a leaf node is reached, upon which time the query will be sent to BE. To exploit QLP, the FE consists of a set of RUs. Each RU independently processes a query popped from the FQQ.

While different RUs can exploit the QLP, processing within each query is sequential due to the inherent nature of depth-first search: the RU would have to finish the current node before deciding whether/how to proceed to the next node in the top-tree. The key challenge in the RU design is thus how to expose intra-query parallelism to improve performance. To that end, we start from a simple hardware design, and gradually introduce architectural optimizations that exploit pipelining.



\paragraph{Baseline Design} Processing a query in the top-tree requires iteratively traversing the top-tree in a DFS manner. We use a stack to maintain the traversal status. Each iteration processes the node at the top of the stack, and pushes the two children nodes back to the stack in the end. Therefore, processing a query consists of the following six stages:

\begin{itemize}[topsep=2pt]
  \item \textbf{FQ}: fetch the query point $Q$ from the FQQ and obtain the query information, including the address of the query stack in the global buffer;
  \item \textbf{RS}: read the top of the stack (TOS) from the query stack; the TOS structure contains the address of the next top-tree node $N$ to be visited;
  \item \textbf{RN}: read the data of $N$ from the global buffer;
  \item \textbf{CD}: calculate the distance $Dist$ between $Q$ and $N$;
  \item \textbf{PI}: push the two child nodes to the stack, and use $Dist$ to decide whether to insert $N$ to the Result Buffer;
  \item \textbf{CL}: issue the query to the BE if a leaf node is reached.
\end{itemize}

The baseline six-stage RU design is illustrated in~\Fig{fig:ru}. The first stage prepares the query data and is required only at the beginning of the query, while the rest five stages are required for each iteration during the query processing. Critically, there is a data dependency between the \textbf{PI} stage that pushes data to the stack and the \textbf{RS} stage that popes data from the stack. This dependency stalls the pipeline for 3 cycles between searching consecutive top-tree nodes. We propose two architectural optimizations that eliminate the stalls and improve performance.


\begin{figure*}[t]
\centering
\includegraphics[width=2\columnwidth]{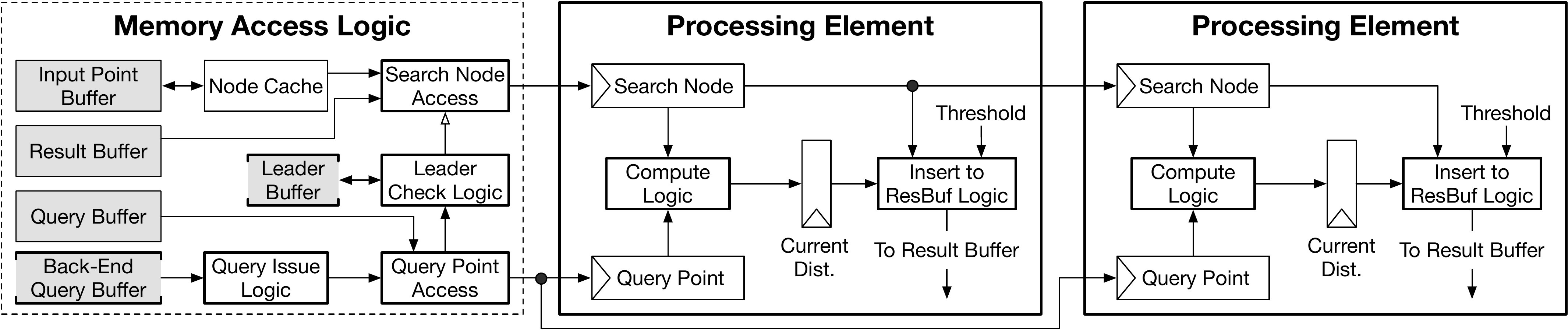}
\caption{Search unit (SU) overview. Each SU has a set of processing elements (PE) operating in the SIMD fashion. The PEs are organized as a 1D systolic array to improve data distribution efficiency. The SU adopts a `` query-stationary'' data-flow where each query is pinned at a PE and the search nodes are streamed through the PEs (i.e., reused by different queries).}
\label{fig:su}
\end{figure*}

\paragraph{Node Forwarding} We observe that the \textbf{PI} stage pushes the current node's two child nodes, $N1$ and $N2$, to the stack; whichever child node gets pushed later will necessarily be popped in the next iteration. Thus, if $N1$ gets pushed first, the \textbf{PI} stage could then directly forward $N2$ to the \textbf{RN} stage, eliminating one stall cycle. To completely remove stalls, we observe that the logic to decide which child node gets pushed first could be determined early in the \textbf{CD} stage rather than waiting until the \textbf{PI} stage. Moving that decision logic earlier to the \textbf{CD} stage completely eliminates stalls.

\paragraph{Node Bypassing} Forwarding eliminates stalls when a node gets to the \textbf{PI} stage. Node bypassing aims at finishing a node early in the pipeline, allowing the next node to start immediately. Specifically, if a node is deemed to be prunable, its entire sub-tree could be skipped (\Sect{sec:algo:ds}). As a result, that node needs not go through the rest steps (i.e., bypassed). Bypassing pruned nodes is particularly significant in NN search when the top-tree is short. A short top-tree would allow the exhaustive search stage to obtain a tighter current nearest distance, exposing more pruned nodes.

To support bypassing, we augment a node's metadata with the distance information, which gets decoded in the \textbf{RN} stage, which in turn generates the bypass signal to let the \textbf{RS} stage start the next node immediately.

\subsection{The Back-End: Search Unit}
\label{sec:arch:su}

Each query that arrives at the BE comes from a particular leaf node in the top-tree. The BE processes the query by exhaustively searching through the leaf node's children, which we call a \textit{Node Set}. The BE exhibits both QLP and NLP. To exploit QLP, the BE incorporates a set of PEs, each handling one query at a time. The Query Point Access logic fetches a query from the Query Buffer at the beginning of query processing, and stores the query point in a PE-local register.~\Fig{fig:su} shows the design of each PE.

Each PE exploits NLP within a query in a pipelined fashion. The PE datapath is pipelined into three stages, through which the nodes in the \textit{Node Set} are streamed. Stage one reads a child node $N$ driven by the Search Node Access logic. Stage two computes the distance $Dist$ between the query point $Q$ and $N$. The final stage decides whether to insert $N$ into the Result Buffer depending on $Dist$. The pipeline is guaranteed to proceed with no stalls because there are no dependencies across different child nodes.


\paragraph{MQMN vs. MQSN} A naive BE design would allow each PE to handle any arbitrary query distributed from the FE to maximize the PE utilization. This design, however, leads to high memory bandwidth requirements because each PE would potentially have to read a different Node List. We call this design Multiple Query Multiple NodeSet (MQMN). The alternative is to force all the PEs to process queries from the same leaf node, which lowers memory bandwidth requirements as different PEs consume the same Node List. We call this design Multiple Query Single NodeSet (MQSN).

While MQSN is memory-efficient, the query issue logic would have to perform an associative search in the BE Query Buffer to find as many queries from the same leaf node as possible. This is key to ensure a high PE array utilization, which, however increases the design complexity and the pipeline cycle time.

\paragraph{Hierarchical MQSN} To enable a complexity-effective PE design while achieving high PE utilization, the BE groups the PEs into several groups, each responsible for only a set of leaf nodes. In this way, the issue logic has a much smaller scheduling window and has fewer PEs to keep occupied, increasing both the scheduling efficiency and the PE utilization.

We call each group a search unit (SU). More specifically, each SU has a set of PEs, a BE Query Buffer that holds queries that are sent to the SU from the FE, a query issue logic to issue queries to the PEs, and a set of address generation logic to access search nodes and query points. With this hierarchical design, we find that MQSN is able to achieve similar PE utilization as MQMN while being complexity-effective. We thus adopt the MQSN design, and will quantify its performance against MQMN later.

The associative search performed by the query issue logic uses the first query in the BQB as the search key, and search the remaining entries in the BQB. The search is done in groups of 32 queries in parallel, and terminates when we find enough queries for the PEs to operate on. The cost of the associative search is amortized across the execution of the found queries, which typically takes two orders of magnitude longer than the associative search.


We find that the overall performance is relatively insensitive to how exactly the leaf nodes are mapped to each SU. Thus, we use a simple policy that uses the low-order bits as the target SU ID. The Query Distribution Network sitting in-between the FE and BE is hard-wired with this logic.

\paragraph{Systolic PE Organization} To reduce the data distribution cost, we organize the different PEs in a SU as a 1D systolic array~\cite{kung1982systolic}.~\Fig{fig:su} shows an example with two PEs, and the nodes in the same Node Set are streamed through the PEs to be reused by different queries. This dataflow is naturally ``query stationary'' as each query is pinned at a PE. Alternatively, the PEs could be organized as a set of SIMD lanes, requiring a data distribution fabric (e.g., bus) with support for multicasts to keep the PEs utilized~\cite{simdvssystolic}.


\paragraph{Approximate Search} Our SU design supports approximate search (\Sect{sec:algo:approx}), which allows a \textit{follower} query to search in the return results of the \textit{leader} queries instead of the leaf node's Node Set. To that end, we augment the memory access logic with a Leader Check logic, which, when determines that the current query could be approximated by a leader, would drive the Search Node Access logic to fetch search nodes from the Result Buffer rather than from the Input Point Buffer. The actual check requires computing the distances between an incoming query and the existing leaders. We reuse the PEs in the SU for these computations.

The leader queries of each leaf node are stored in a local Leader Buffer, which we cap at 16 entries guided by the profiling results on the KITTI dataset~\cite{geiger2012we}. The leader group stops growing after the buffer is full, which we find rare in our experiments, to simplify the hardware implementation. It is worth noting that capping the Leader Buffer \textit{improves} accuracy because more queries will be searched exactly without relying on the leaders.


\paragraph{Node Cache} While the MQSN design significantly reduces the Node Set load traffic, loading from the Node Sets still contributes to over half of the total memory traffic. We observe that queries consecutively issued from the FE are likely from a small set of leaf nodes. We propose a cache design to capture the locality when loading the Node Sets to further reduce the memory traffic.

The node cache is organized as a set of entires, each containing the nodes in one Node Set. The nodes in each entry is organized as a FIFO queue because nodes in a Node Set are accessed sequentially. While there is a need to associatively search different entries to determine whether a particular Node Set is in the cache, the nodes within an entry could be accessed as a FIFO, greatly simplifying the hardware implementation.

%% file: eval.tex
\section{Evaluation}
\label{sec:eval}

We first describe the evaluation methodology (\Sect{sec:eval:exp}). We then analyze the area of the \proj accelerator (\Sect{sec:eval:area}). We then compare the performance and energy consumption of the accelerator over the baseline system for both the KD-tree search alone and the end-to-end pipeline (\Sect{sec:eval:res}). We tease apart the contributions from different optimizations (\Sect{sec:eval:opt}), and analyze how the performance of \proj is sensitive to different resource configurations (\Sect{sec:eval:sen}).


\subsection{Experimental Methodology}
\label{sec:eval:exp}

\paragraph{Hardware Implementation} We synthesize, place, and route the accelerator datapath using Synposys and Cadence tools in a 16nm process technology, with memories generated using an SRAM compiler. Power is estimated using Synopsys PrimeTimePX by annotating the switching activity. The datapath is able to be clocked at 500~MHz. The DRAM energy is estimated using Micron's DDR4 specification~\cite{MicronDDR4} and power calculator~\cite{micronddr4powercalc}. We then use a cycle-accurate simulator parameterized with the synthesis and memory estimations to drive the performance and energy analysis.

\paragraph{Dataset} We evaluate \proj on the widely-used KITTI Odometry dataset~\cite{geiger2012we}. We use the first 11 sequences in the dataset that has ground truth. Each sequence consists of hundreds to thousands of point cloud frames. The point cloud in the KITTI dataset is obtained using the popular Velodyne HDL-64E LiDAR~\cite{VelodyneHDL64E}, representative of today's point cloud acquisition system. We report the average results across all the frames, unless noted otherwise.

\paragraph{Metrics} We evaluate \proj in performance, energy, and accuracy. We show both the KD-tree time and the end-to-end registration time for all the frames in the entire sequence. The accuracy is measured using standard rotational and translational errors~\cite{geiger2012we}.

\paragraph{Baseline} While the performance characterizations are performed on a CPU-based implementation~(\Sect{sec:char:dse}) as most of today's point cloud registration pipelines are implemented on the CPU~\cite{kitti_odometry}, for a fair evaluation we use a GPU/CUDA implementation of KD-tree search from the popular FLANN library~\cite{muja2014scalable}. KD-tree search on the GPU is about 8--20$\times$ faster than on the CPU.

We use a CPU-GPU setup as the baseline system. The KD-tree searches run on the GPU while all other operations run on the CPU. The CPU is a 32-core Xeon Silver 4110 Processor, and the GPU is an Nvidia GeForce RTX 2080 Ti. We use the widely-used Point Cloud Library (PCL)~\cite{rusu2011point} to develop the registration pipelines, and integrate the FLANN's implementation of KD-tree search. The GPU power is measured at 100~Hz using the \texttt{nvidia-smi} utility, and the CPU power is measured using the Intel RAPL energy counters~\cite{david2010rapl} via directly reading the processor MSRs~\cite{xeondatasheetv2}.



To demonstrate the generally applicability of \proj, we evaluate it on two Pareto-optimal designs of the point cloud registration pipeline (\Fig{fig:dse} in \Sect{sec:char:dse}): \texttt{DP4} that optimizes for performance and \texttt{DP7} that optimizes for accuracy.


\subsection{Area Analysis}
\label{sec:eval:area}

We configure the \proj accelerator to have 64 RUs, 32 SUs, and 32 PEs per SU. We size the on-chip SRAM to accommodate about 130,000 points per frame, representative of the point cloud density acquired in the real-world. In particular, the Input Point Buffer and the Query Buffer are both sized at 1.5~MB; the Query Stack Buffer is sized at 1.2~MB, accommodating a maximal top-tree height of 18, sufficient for the KITTI dataset; the FE Query Queue is sized at 1.5~MB, and the BE Query Buffer is sized at 1~KB per SU, holding 128 BE queries at a time. The Node Cache is configured at 128~KB. Finally, the Result Buffer is set at 3~MB, which is double-buffered to interface with the DRAM to take the result write traffic off the critical path. Overall, the SRAM is estimated to take \SI{8.38}{\mm\squared}.

The datapath area of each RU and each SU's PE is mostly dominated by the logic that computes the euclidean distance between two points using 32-bit floating point arithmetics. The total combinational logic occupies about \SI{7.19}{\mm\squared}. Overall, 53.8\% of area is taken by SRAM and 46.2\% is occupied by compute logic.


\subsection{Performance and Power Comparisons}
\label{sec:eval:res}

\begin{figure}[t]
    \vspace{-10pt}
	\centering
    \subfloat[Accuracy-oriented \texttt{DP7}.]{
      \label{fig:pp_dp7}
      \includegraphics[width=0.48\columnwidth]{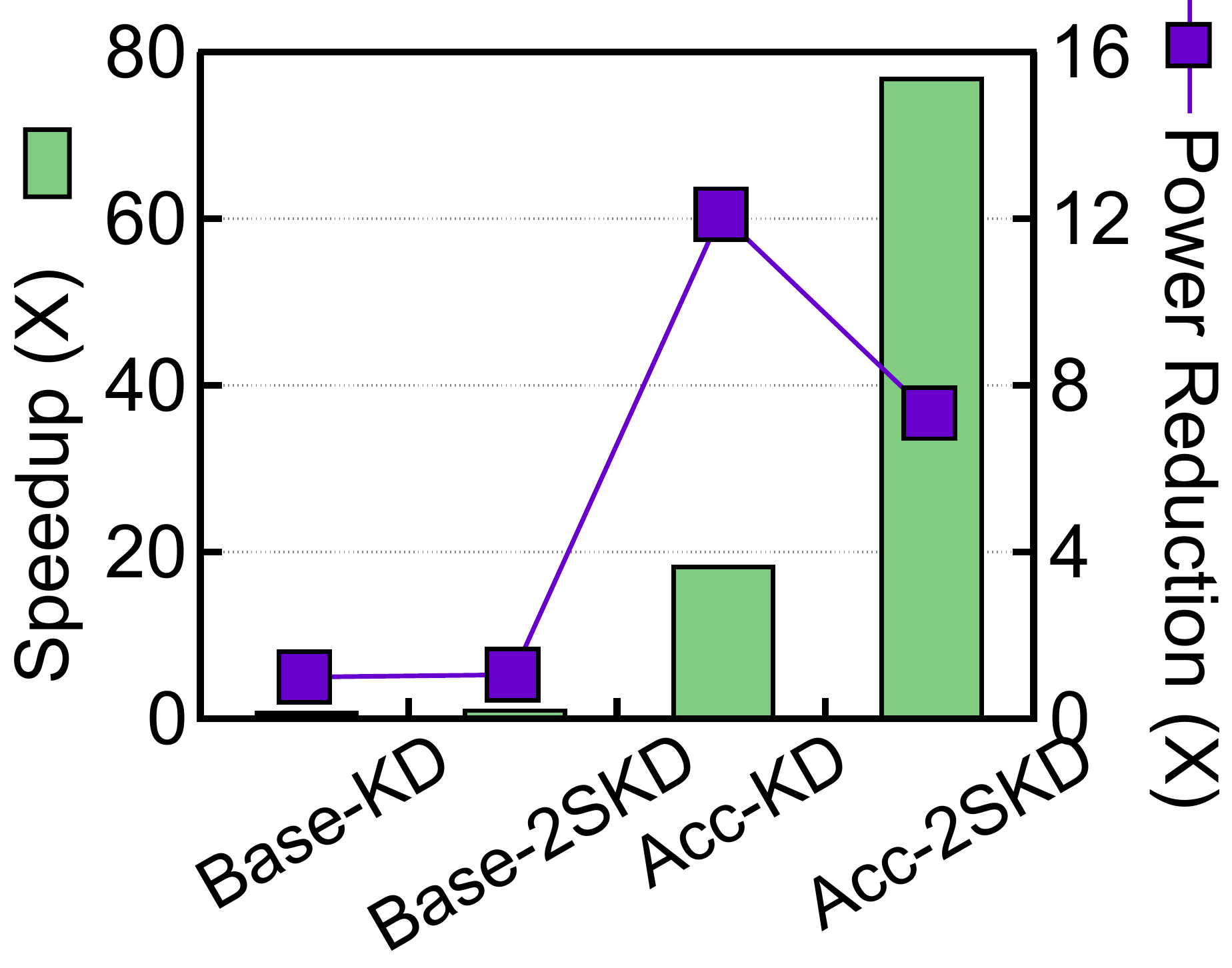}
    }
    \subfloat[Performance-oriented \texttt{DP4}.]{
      \label{fig:pp_dp4} 
      \includegraphics[width=0.48\columnwidth]{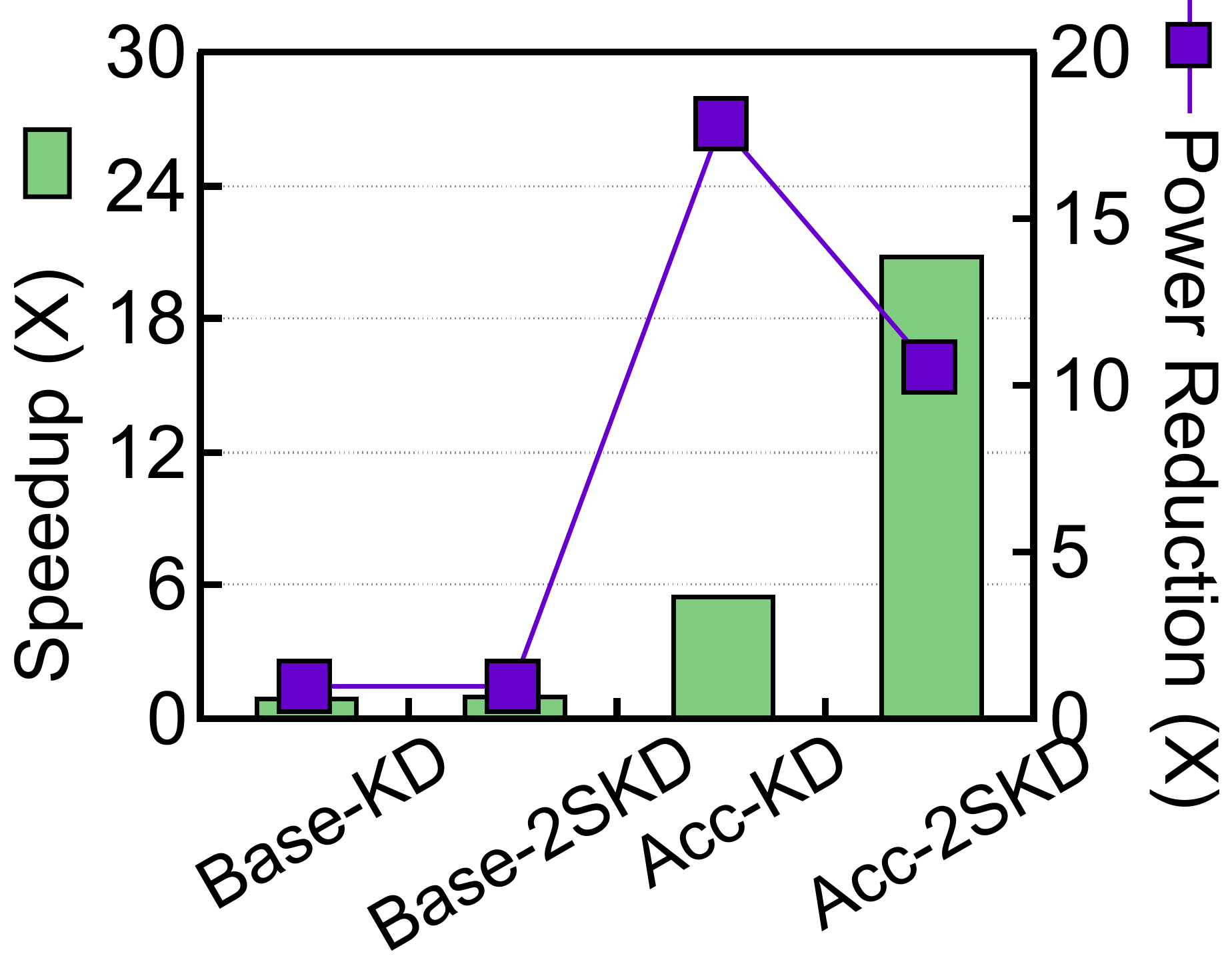}
    }
    \vspace{-5pt}
    \caption{KD-tree search speedup and power reduction on two Pareto-optimal designs: \texttt{DP7} is accuracy-oriented and \texttt{DP4} is performance-oriented.}
    \label{fig:pp} 
\end{figure}

\paragraph{Speedup} \proj achieves significant speedup in KD-tree search compared to the baseline. Using the accuracy-oriented design point \texttt{DP7} as an example,~\Fig{fig:pp_dp7} shows the KD-tree search speedup of the \proj accelerator running both the original KD-tree (\sys{Acc-KD}) and the two-stage KD-tree with a top-tree height of 10 (leaf-set size of about 128) (\sys{Acc-2SKD}) compared to the GPU baseline that runs the original KD-tree, i.e., leaf-set size 1 (\sys{Base-KD}). For comparison purposes, we also show the speedup of the GPU running the two-stage KD-tree with the same top-tree height 10 (\sys{Base-2SKD}). Note that both \sys{Acc-KD} and \sys{Acc-2SKD} do \textit{not} apply approximate search here, i.e., no accuracy loss.

\sys{Acc-2SKD} achieves 77.2$\times$ speedup in KD-tree search compared to \sys{Base-2SKD}, which in turn is 28.3\% faster than \sys{Base-KD}. \sys{Acc-KD} however, is ``only'' 18.7$\times$ faster than the \sys{Base-KD} baseline. This is because using the original KD-tree, the accelerator's performance is almost completely bottlenecked by the recursive search in the top-tree while the back-end SUs are almost always idle, leading to resource under-utilization. This confirms the need to co-design accelerator with the new data structure that exposes parallelism. Compared to the CPU implementation of KD-tree (not shown), \sys{Acc-2SKD} achieves a speed up of 392.2$\times$.

The speedup on KD-tree search translates to significant end-to-end performance improvement. Specifically, \sys{Acc-2SKD} reduces the overall registration time by 41.7\% and 86.6\% compared to the GPU baseline \sys{Base-KD} and the CPU-only implementation, respectively.


\proj also achieves high speedups in the performance-oriented design point \texttt{DP4}.~\Fig{fig:pp_dp4} shows the performance comparisons across different systems for \texttt{DP4}. \sys{Acc-2SKD} achieves about 21.0$\times$ speedup compared to \sys{Base-2SKD} on KD-tree search alone, which translates to about 13.6\% end-to-end performance improvement. The speedup on the \texttt{DP4} is lower than \texttt{DP7} because in optimizing for performance \texttt{DP4} uses tight search criteria that leads to much fewer exhaustive searches. For instance, the Normal Estimation stage in \texttt{DP4} uses a radius of 0.30 while using a radius of 0.75 in \texttt{DP7}. A relaxed radius exposes more exhaustive searches, which could benefit from the SU design of \proj. Overall, the performance improvements on two very different design points demonstrate the general applicability of \proj.

\paragraph{Power Reductions} We overlay the power reductions on the right $y$-axis
in~\Fig{fig:pp_dp7} and~\Fig{fig:pp_dp4}. \sys{Acc-2SKD} achieves about 7$\times$ and 10.5$\times$ power reductions compared to \sys{Base-KD} on KD-tree search for \texttt{DP7} and \texttt{DP4}, respectively. The reduction along with the speedup further translates to significant energy savings (i.e., power-efficiency). For instance, \sys{Acc-2SKD} reduces the energy consumption of \sys{Base-KD} by a factor of 220.2 on \texttt{DP4}. Breaking down the energy consumption of \texttt{DP4}, the PE contributes to about 53.7\% of the total energy consumption. The rest of energy is contributed by SRAM read(34.8\%), SRAM write(8.0\%), Leakage(3.3\%), and DRAM read/write(0.2\%). Over the end-to-end pipeline, \sys{Acc-2SKD} achieves about a 3.0$\times$ power reduction compared to \sys{Base-KD}.

The power consumption of \sys{Acc-KD} is lower than \sys{Acc-2SKD}, because \sys{Acc-KD} does not expose exhaustive searches in the leaf nodes, and thus exclusively exercises the RUs while leaving the SUs idle. It trades lower power for lower performance. As a result, its overall energy consumption is about 2.5$\times$ higher than \sys{Acc-2SKD}.

\paragraph{Approximate Search} We empirically choose 1.2 meters as the approximate threshold (\Sect{sec:algo:approx}) for the NN search, and use 40\% of the original radius as the threshold in the radius search. Using these settings, the approximate search has no impact on translation errors, and 
increases the rotational error only by \SI{0.05}{\degree}/meter on \texttt{DP4} and \SI{0.0006}{\degree}/meter on \texttt{DP7}.

Using \texttt{DP7} as an example, the approximate KD-tree search achieves about 11.1$\times$ performance improvements over \sys{Acc-2SKD}, translating to 7.5\% end-to-end performance improvement. The improvement is a direct result of the compute reduction: the approximate algorithm reduces the number of nodes visited during search by 72.8\%, to which NN search contributes 41.6\% and radius search contributes 31.2\%.


\begin{figure}[t]
  \begin{minipage}[t]{0.47\columnwidth}
    \centering
    \includegraphics[trim=0 0 0 0, clip, height=1.3in]{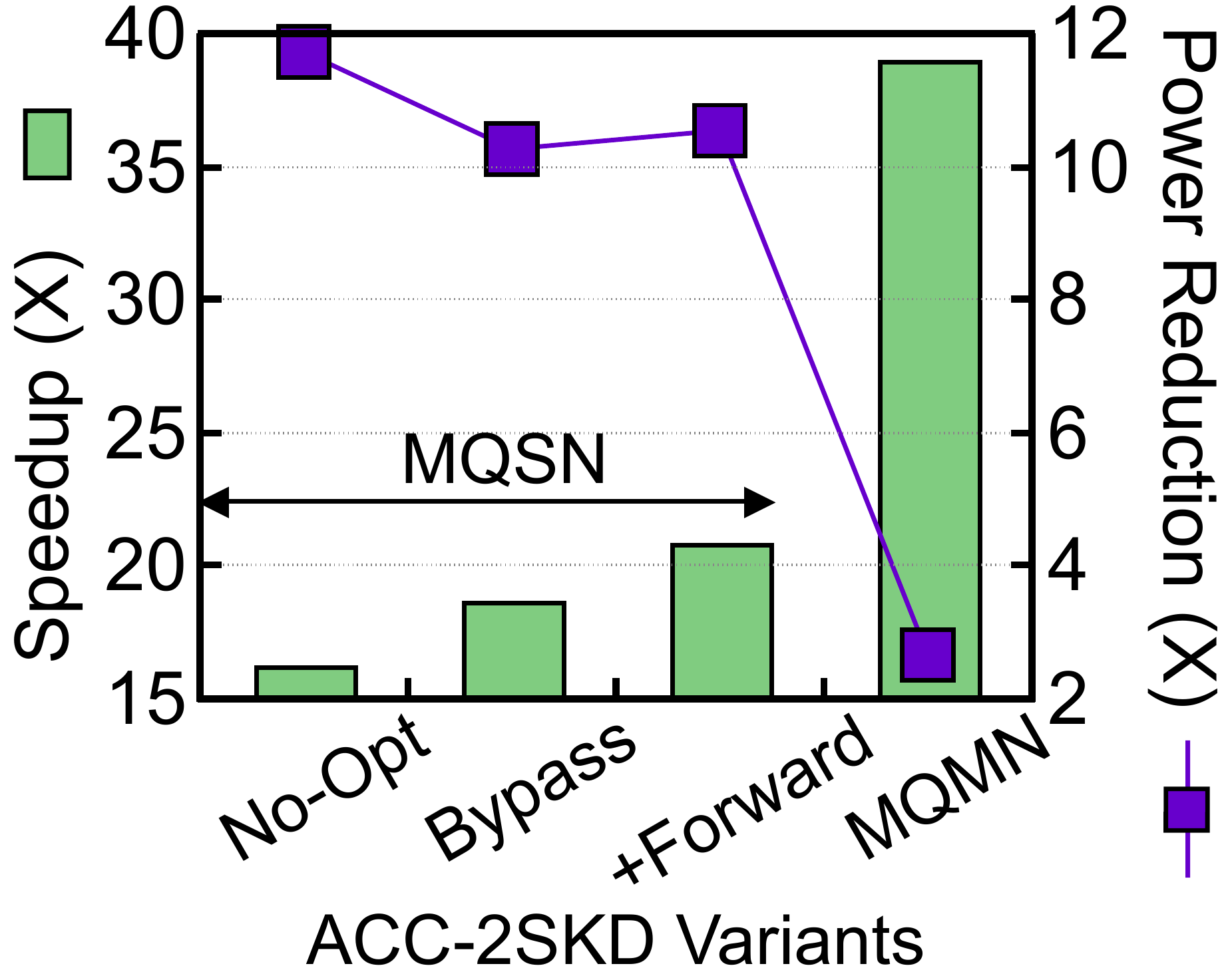}
    \caption{The speedup and power reduction of architectural optimizations.}
    \label{fig:opt}
  \end{minipage}
  \hfill
  \begin{minipage}[t]{0.47\columnwidth}
    \centering
    \includegraphics[trim=0 0 0 0, clip, height=1.3in]{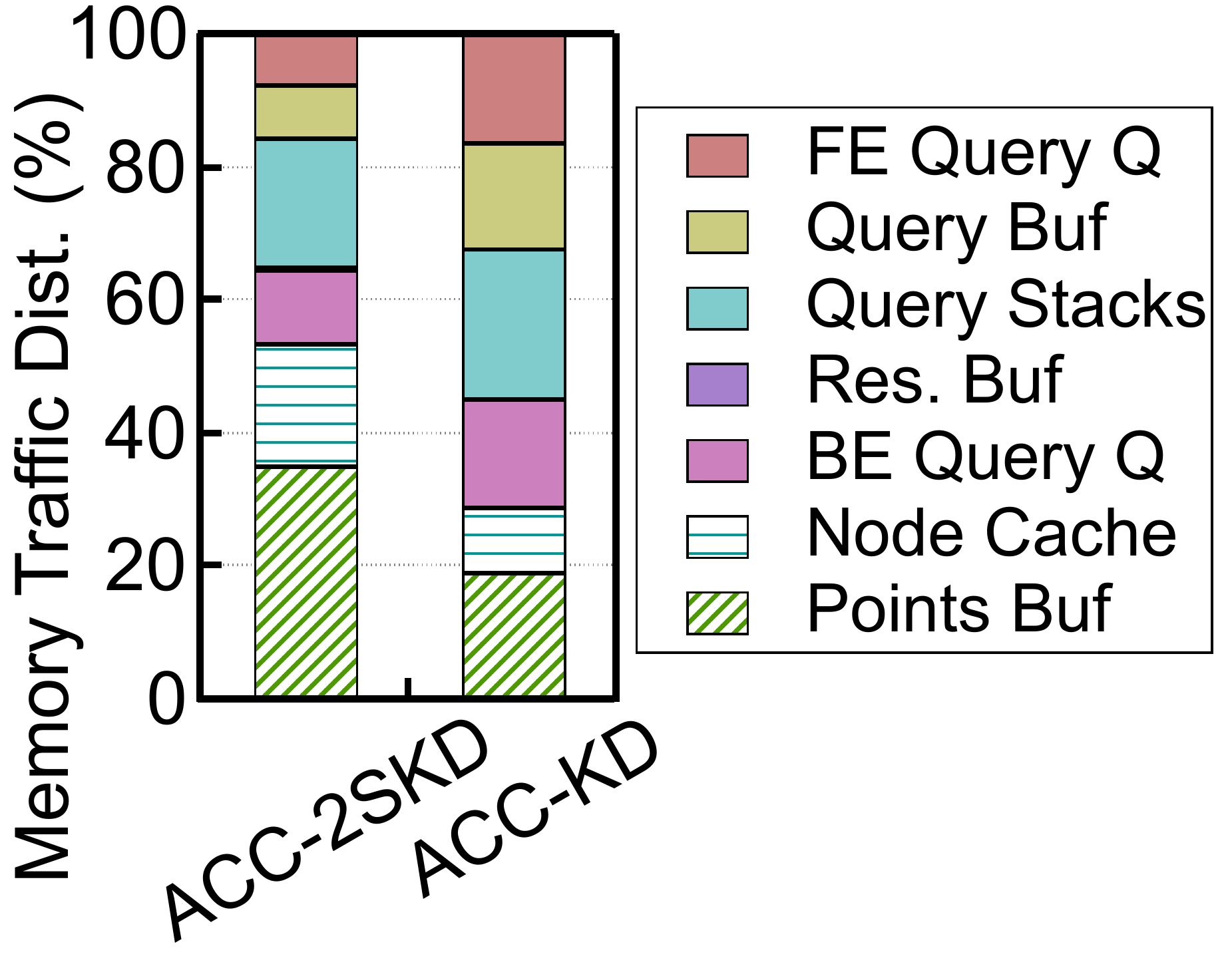}
    \caption{The memory traffic distributions. Point cache alleviates Point buffer traffic.}
    \label{fig:memtraffic_dist}
  \end{minipage}
\end{figure}

\subsection{Optimization Effects}
\label{sec:eval:opt}

\paragraph{Bypassing and Forwarding} In the RU design, bypassing allows pruned nodes to take an early exit from the pipeline, and forwarding allows the node that will soon be at the top of the query stack to start immediately. Both techniques help reduce pipeline stalls and improve the performance.~\Fig{fig:opt} shows the speedup over \sys{Base-KD} of three \sys{Acc-2SKD} variants: without either technique (\texttt{No-Opt}), with just bypassing (\texttt{Bypass}), and with both bypassing and forwarding (\texttt{+Forward}). Bypassing improves the performance by about 13.1\%; forwarding further achieves 10.5\% improvements.

\paragraph{MQMN vs. MQSN} MQMN allows different PEs from the same SU to process queries from different leaf nodes at the cost of additional memory traffics (\Sect{sec:arch:su}).~\Fig{fig:opt} shows the speedup of the MQMN organization of \sys{Acc-2SKD} over \sys{Base-KD}, and compares it against MQSN variants.MQMN doubles the performance of the best MQSN variant (\texttt{+Forward}). However, the additional memory traffic significantly increases the power consumption. The right $y$-axis in~\Fig{fig:opt} overlays the power reductions of various schemes over \sys{Base-KD}. MQSN's power consumption is almost 4$\times$ worse than \texttt{+Forward}, leading to 2$\times$ energy.

\paragraph{Node Cache} Node Cache reduces the global Points Buffer traffic and thus saves energy.~\Fig{fig:memtraffic_dist} shows the memory traffic distributions across different data structures. In \sys{ACC-2SKD}, the Points Buffer traffics would account for 53\% of the total traffics without the Node Cache, and are reduced to 35\% with the Node Cache. By directing 18\% of the memory traffic to a smaller memory, the Node Cache reduces the energy by 5.9\% (not shown). \sys{ACC-KD} has very few exhaustive searches and thus much lower Points Buffer traffics. As a result, the Points Buffer traffics contribute to only 29\% of the total traffic; the effect of the Node Cache is smaller.

\subsection{Sensitivity Analysis}
\label{sec:eval:sen}

We study how the performance and energy of \proj vary with hardware resources and software parameters.

\begin{figure}[t]
    \vspace{-5pt}
	\centering 
    \subfloat[Performance vs. power.]{
      \label{fig:sweep_perf_power}
      \includegraphics[height=1.35in]{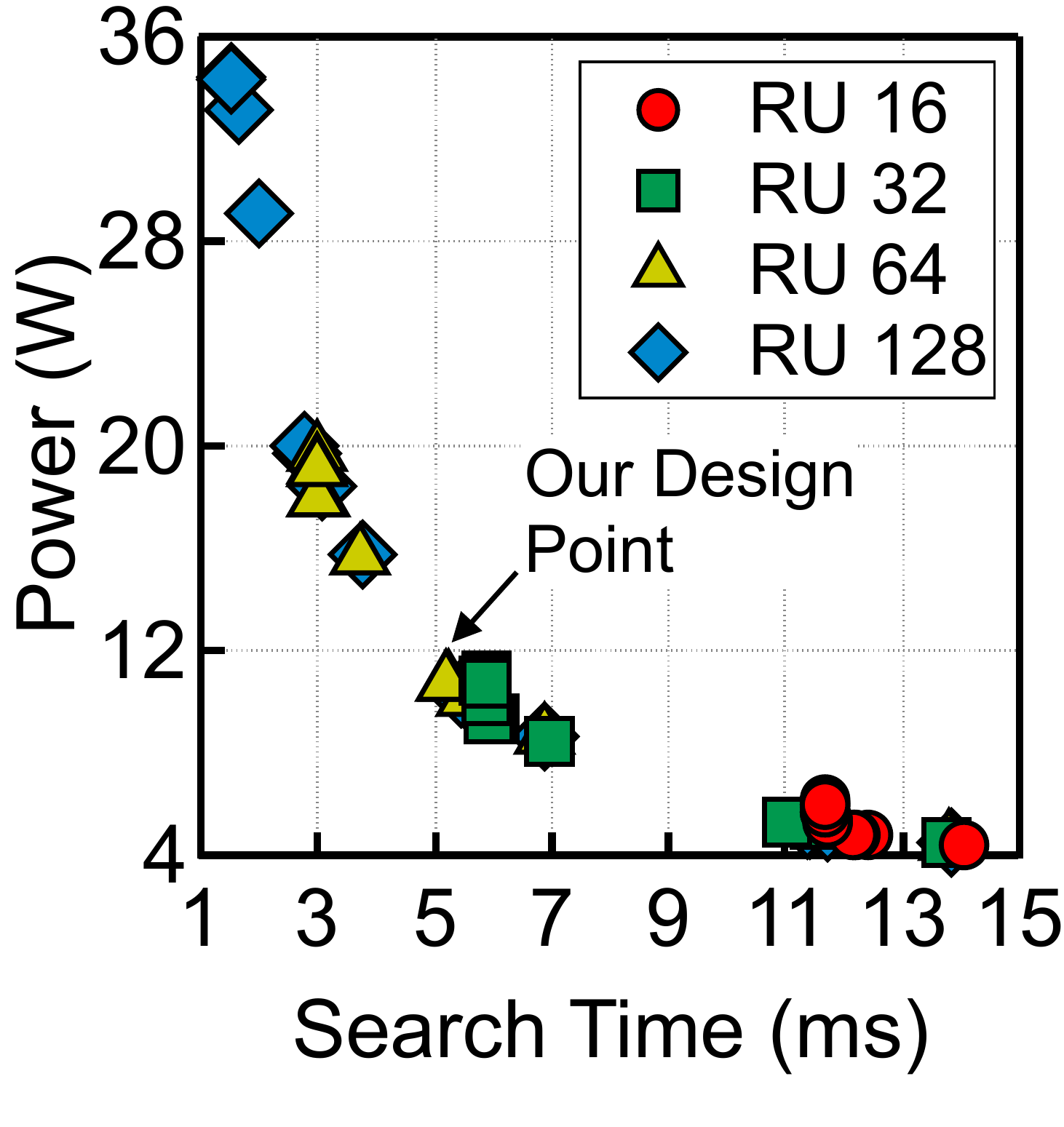}
    }
    \subfloat[Performance comparison.]{
      \label{fig:sweep_perf} 
      \includegraphics[height=1.35in]{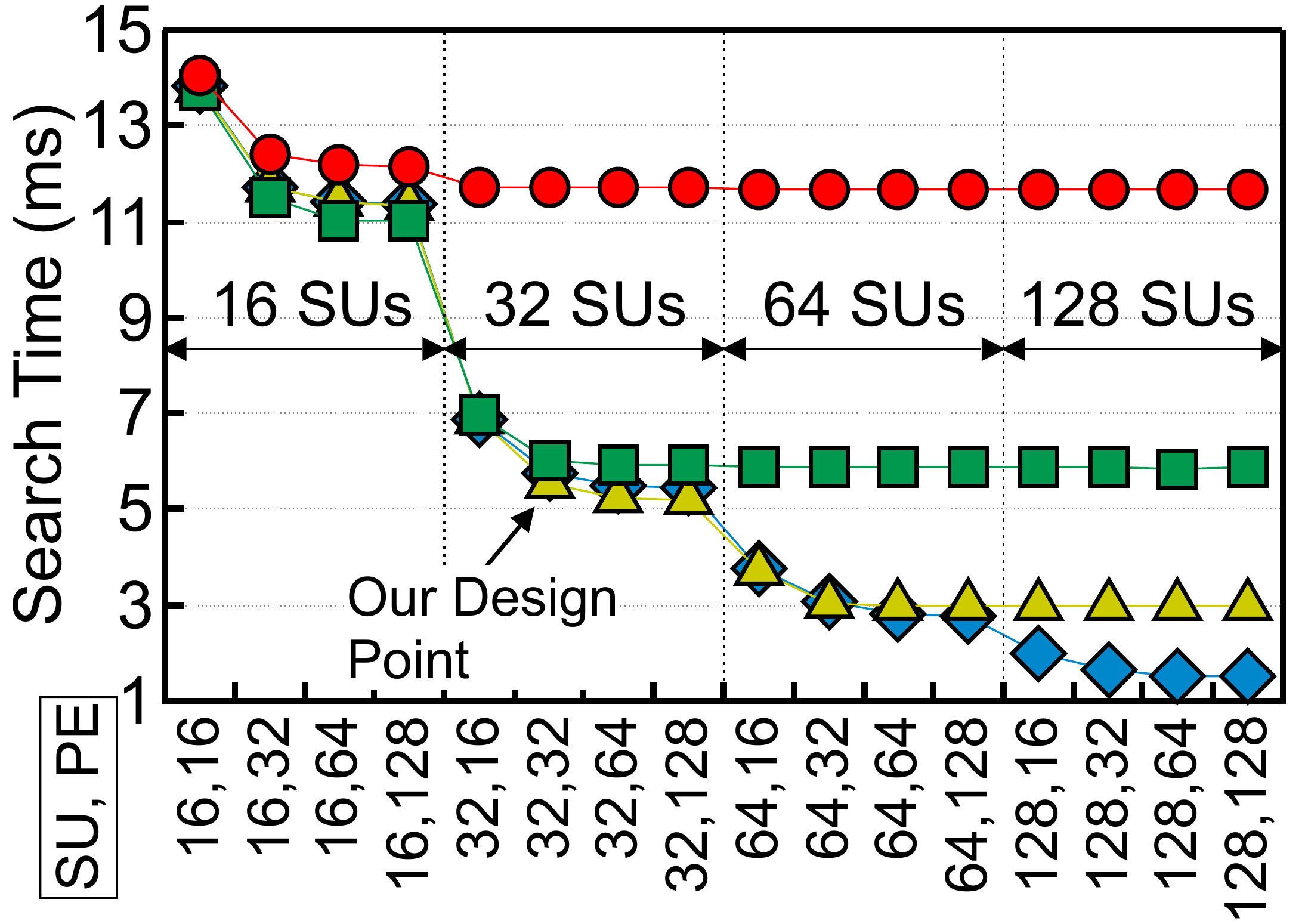}
    }
    \vspace{-5pt}
    \caption{Performance and power sensitivity to three hardware parameters: number of RUs, number of SUs, and number of PEs per SU. Both figures share the same legend.}
    \label{fig:sweep}
\end{figure}

\paragraph{Hardware Configurations} We study three key parameters: the number of RU, the number of SU, and the number of PEs per SU. We sweep all three parameters from 16, 32, 64, through 128.~\Fig{fig:sweep_perf_power} shows the KD-tree search time and power under all 64 configurations. Overall, as performance improves the power consumption also increases.~\Fig{fig:sweep_perf} shows a detailed performance comparison of different configurations, where different curves represent different RU counts and the $x$-axis sweeps the SU and PE counts.

When the RU count is low, e.g., 16 and 32, the performance is bottlenecked by the front-end. Thus, improving the back-end capabilities by increasing the SU and PE counts improves the overall speed only marginally. As the RU count increases to 64, the accelerator becomes balanced. Our design choice of 64 RUs, 32 RUs, and 32 PEs per SU sits on the ``knee of the curve'', indicating a complexity-efficient design decision.

\begin{figure}[h]
\vspace{-5pt}
\centering
\includegraphics[width=.9\columnwidth]{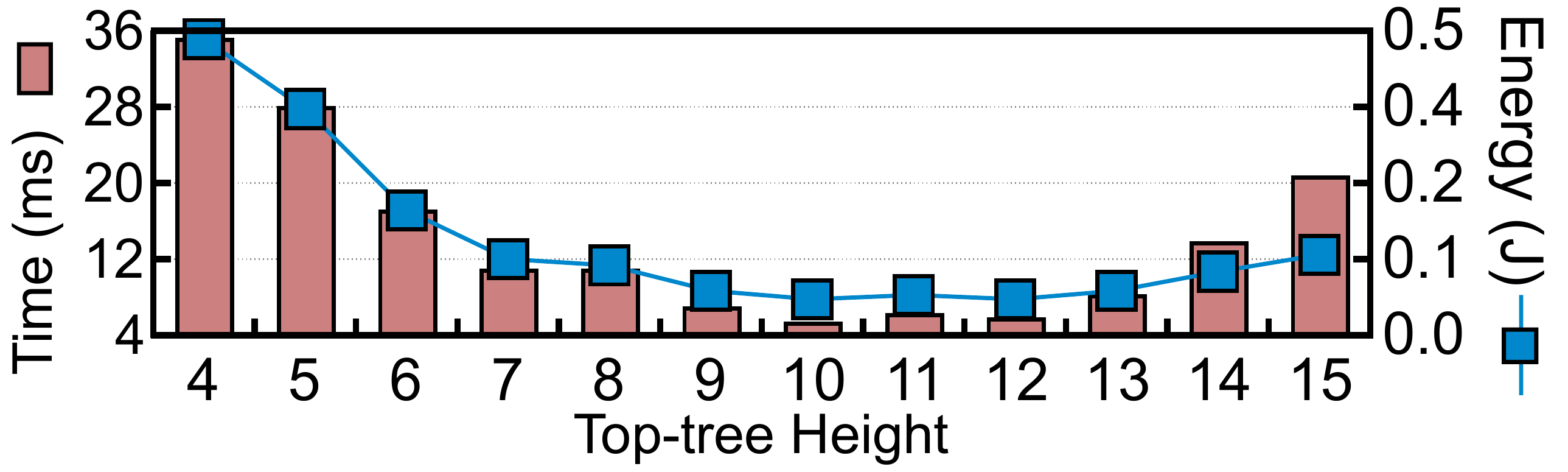}
\vspace{-5pt}
\caption{Search time and energy vary with top-tree height.}
\label{fig:sweep_height}
\vspace{-5pt}
\end{figure}

\paragraph{Software Configurations} The results we have shown so far assume a top-tree height of 10 (about 128 children per leaf node in the top-tree).~\Fig{fig:sweep_height} shows the KD-tree search time and energy consumption as a function of the top-tree height. The performance initially increases as the top-tree height increase. This is because higher top-trees have less redundancy in exhaustive searches. However, the performance reaches a diminishing return when the top-tree height reaches around 10, beyond which the performance decreases. This is because a high top-tree requires more recursive search in the RU, thus reducing the node-level parallelism within each query that could be exploited by the SUs.

We find that the optimal top-tree height (10) is largely consistent across different KD-tree search instances in the pipeline. The optimal top-tree height mainly depends on two factors: 1) the points in a data frame, and 2) the hardware organization. Given a specific registration pipeline, different KD-tree search instances share these two factors and thus share the same optimal top-tree height.

%% file: related.tex
\section{Related Work}
\label{sec:related}

\paragraph{Point Cloud Registration} Point cloud registration pipelines generally fall under three categories depending on the density of points that are used for registration. On one extreme is algorithms that use all the points for registration~\cite{besl1992method, behley2018efficient, park2017probabilistic}, which tolerate outliers but are computationally prohibitive in real-time. On the other extreme are algorithms that use only (sampled) feature points~\cite{Jiang2009Registration, deschaud2018imls}, which are efficient in compute, but could suffer from local minima. 

The point cloud registration pipeline studied in this paper represents a trade-off between the two extremes, and is the predominant choice today~\cite{Xu2017An, Choi_2015_CVPR, huang2018coarse, yang2016fast, zhang2014loam}. It uses feature points for coarse-grained, initial estimation while using all the points for fine-tuning. While prior work proposes specific design points that make specific accuracy-speed trade-offs, we construct a flexible pipeline that let us perform design space exploration, which reveals Pareto-optimal design points that drive our performance bottleneck analysis.


Recent work has also using Deep Neural Networks (DNN) for point cloud registration. End-to-end DNNs are susceptible to and are limited to specific registration cases such as pose estimation~\cite{wang2019densefusion}. DNNs are mostly used to replace certain stages of the registration pipeline such as key point detection~\cite{qi2017pointnet++, feng20163d}, normal estimation~\cite{boulch2016deep}, description calculation~\cite{elbaz20173d}, and fine-tuned ICP~\cite{liu2006three} while relying on the overall pipeline architecture as we described in~\Sect{sec:char:pipe}.

To our best knowledge, this is also the first paper that proposes hardware accelerator for point cloud registrations while prior work mostly focuses on algorithmic developments.



\paragraph{KD-Tree Search Acceleration} KD-tree search is widely used in application domains beyond point cloud registration, such as graphics~\cite{horn2007interactive, popov2007stackless}, data analytics~\cite{xiao2010efficient, ooi1987spatial}, and image/video processing~\cite{xu2009efficient, huang2010video}. \proj accelerates the fundamental KD-tree search algorithm, and is applicable to these application domains as well.

Accelerating KD-tree search has been mostly explored in the context of Map-Reduce~\cite{aly2011distributed}, GPU~\cite{Qiu2009GPU, Gieseke2014Buffer, Heinzle2008A, Kuhara2013An}, and FPGA~\cite{Winterstein2013FPGA}. The \proj accelerator differs from prior attempts in its systematic and comprehensive exploitation of different forms of parallelism in KD-tree search. Specifically, our \proj accelerator exploits query-level parallelism (QLP) and node-level parallelism (NLP) both in the top-tree traversal and in the exhaustive searches. Most prior work exploits only QLP without NLP~\cite{aly2011distributed, Qiu2009GPU, Heinzle2008A, Kuhara2013An}. Buffer KD-tree~\cite{Gieseke2014Buffer} allows for NLP in the leaf nodes, but does not permit NLP in tree traversal. Heinzle et al.~\cite{Heinzle2008A} exposes NLP in tree traversal, but does not exposes NLP in leaf nodes. Our accelerator design also incorporates a set of architectural mechanisms (e.g., node forwarding/bypassing, MQSN, systolic PE organization) that are unobtainable in general-purpose hardware such as GPUs.





\paragraph{Approximate KD-Tree Search} The approximate nature of many robotics and graphics applications that require neighbor information has spurred much interest in approximate KD-tree/KNN search algorithms~\cite{arya1998optimal, miclet1983approximative, greenspan2003approximate, ma2002low, purcell2005photon, Heinzle2008A}. Our approximate KD-tree search algorithm differs from prior work in two key ways. First, we quantify the extent to which KD-tree search can be approximated in the context of end-to-end registration accuracy while prior work mostly focuses on the accuracy of KD-tree search alone. Second, our approximate search algorithm applies to both NN search and radius search while most prior work is limited to NN search.





%% file: conc.tex
\section{Conclusion}
\label{sec:conc}

With the proliferation of 3D sensors and the rising need for ubiquitous 3D perception, point cloud processing is increasingly becoming the cornerstone of many machine perception applications, and architects must be ready for that. To our best knowledge, this is the first paper that comprehensively characterizes and addresses the performance bottlenecks of point cloud registration. The key to our approach is to co-design the data structure, algorithm, and the accelerator of the key compute kernel. Our work provides the first answer, not the final answer, in a promising direction of research.